\pdfoutput=1
\documentclass[11pt]{article}

\usepackage[final]{acl}

\usepackage{times}
\usepackage{latexsym}

\usepackage[T1]{fontenc}
\usepackage[utf8]{inputenc}

\usepackage{microtype}

\usepackage{inconsolata}

\usepackage{graphicx}

\usepackage{algorithm}
\usepackage{algpseudocode}
\usepackage{amsmath}
\usepackage{bbm}
\usepackage{multirow}
\usepackage{booktabs}
\usepackage{colortbl}
\usepackage{amssymb}
\usepackage{tabularx}
\usepackage{xcolor}
\usepackage{hyperref}
\usepackage[most]{tcolorbox}
\title{Gumbel Reranking: Differentiable End-to-End Reranker Optimization}

  \author{
      Siyuan Huang\textsuperscript{1,3,\footnotemark[1]},
      Zhiyuan Ma\textsuperscript{2},
      Jintao Du\textsuperscript{2},
    Changhua Meng\textsuperscript{2},
    Weiqiang Wang\textsuperscript{2},\\
      \textbf{
      Jingwen Leng\textsuperscript{4},
      Minyi Guo\textsuperscript{4},
          Zhouhan Lin \textsuperscript{1,\footnotemark[2]}} \\
  $^{1}$LUMIA Lab, Shanghai Jiao Tong University, \\
  $^{2}$Tiansuan Lab, Ant Group Co., Ltd.\\
  $^{3}$SJTU Paris Elite Institute of Technology \\
  $^{4}$Shanghai Jiao Tong University \\
\texttt{siyuan\_huang\_sjtu@outlook.com}, \texttt{lin.zhouhan@gmail.com}\\ \texttt{\{mazhiyuan.mzy,lingke.djt,changhua.mch,weiqiang.wwq\}@antgroup.com}\\}

\begin{document}
\maketitle

\renewcommand{\thefootnote}{\fnsymbol{footnote}} %将脚注符号设置为fnsymbol类型，即特殊符号表示
\footnotetext[1]{Work done during an internship at Ant Group.}
\footnotetext[2]{Corresponding Author.}

\begin{abstract}
RAG systems rely on rerankers to identify relevant documents. However, fine-tuning these models remains challenging due to the scarcity of annotated query-document pairs. Existing distillation-based approaches suffer from training-inference misalignment and fail to capture interdependencies among candidate documents. To overcome these limitations, we reframe the reranking process as an attention-mask problem and propose Gumbel Reranking, an end-to-end training framework for rerankers aimed at minimizing the training-inference gap. In our approach, reranker optimization is reformulated as learning a stochastic, document-wise Top-$k$ attention mask using the \textit{Gumbel trick} and \textit{relaxed top-$k$ sampling}. This formulation enables end-to-end optimization by minimizing the overall language loss. Experiments across various settings consistently demonstrate performance gains, including a 10.4\% improvement in recall on HotpotQA for distinguishing indirectly relevant documents.

\footnotetext[3]{Code is available at this \href{https://github.com/LUMIA-Group/Gumbel-Reranking}{url}.}
\end{abstract}

\section{Introduction}

\begin{figure*}[t]
\centering
  \includegraphics[width=\linewidth]{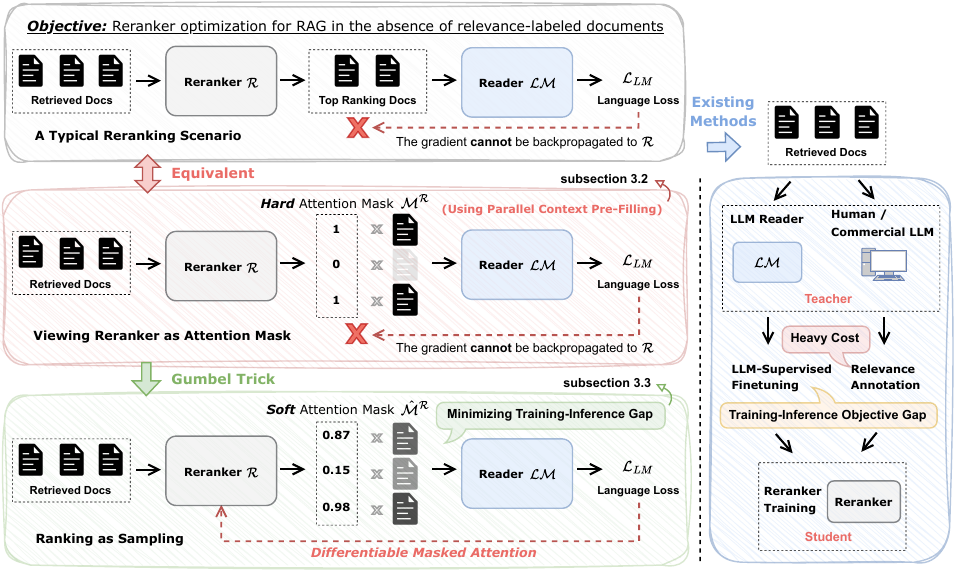}
\caption{Vanilla reranker training methods for RAG systems typically rely on supervised learning of query-document pairs, which is limited by the scarcity of labeled data. To address this issue, existing methods leverage various LLM-supervised losses. However, this can lead to potential gaps between training and inference. In contrast, G-Rerank frames reranker training as learning a stochastic, document-wise top-\(k\) attention mask. This enables end-to-end optimization by minimizing language loss, ensuring better alignment between training and inference.}
  \label{fig:introduction}
\end{figure*}

Retrieval-Augmented Generation (RAG) has shown great potential in natural language processing tasks~\cite{DBLP:conf/nips/LewisPPPKGKLYR020, DBLP:conf/icml/GuuLTPC20, DBLP:conf/eacl/IzacardG21, DBLP:conf/icml/BorgeaudMHCRM0L22}. Despite their remarkable progress, retrieval models in RAG systems—comprising both the retriever and reranker—are typically trained on publicly available datasets and often struggle with long-tail queries requiring domain-specific knowledge.  As a result, they necessitate further fine-tuning for specific downstream tasks~\cite{DBLP:conf/naacl/GlassRCNCG22, DBLP:conf/naacl/ShiMYS0LZY24}. A key challenge in this context is the scarcity of labeled query-document pairs~\cite{DBLP:conf/acl/LeeCT19, DBLP:journals/tacl/SachanLYZPZ23}. Therefore, a critical research question is how to end-to-end optimize the retrieval models of RAG systems solely relying on the system’s final language modeling loss.

Recent efforts to improve retriever or reranker in RAG systems have explored distilling knowledge from LLMs into retrieval components. Techniques such as attention-based distillation~\cite{DBLP:conf/iclr/IzacardG21} and perplexity-based distillation~\cite{DBLP:conf/nips/SachanRHDY21, DBLP:conf/naacl/ShiMYS0LZY24, DBLP:conf/iclr/Lin0CSL00KSLZY24, DBLP:journals/jmlr/IzacardLLHPSDJRG23, DBLP:conf/naacl/GlassRCNCG22} have yielded notable performance gains. However, these methods still exhibit critical limitations. First, although these methods claim to be end-to-end optimized, they focus on LLM-supervised losses like KL divergence~\cite{DBLP:journals/jmlr/IzacardLLHPSDJRG23, DBLP:conf/naacl/GlassRCNCG22} or marginalization~\cite{DBLP:conf/nips/SachanRHDY21, DBLP:conf/naacl/ShiMYS0LZY24, DBLP:conf/iclr/Lin0CSL00KSLZY24}, which do not directly minimize the RAG system’s final generation loss, leading to potential misalignment between training and evaluation objectives. Additionally, attention-based distillation suffers from the distraction problem, where accumulated attention scores do not always reflect document relevance~\cite{DBLP:conf/acl/KeK00MB24, DBLP:journals/corr/abs-2408-11745}. While perplexity-based distillation methods evaluate each candidate document in isolation, neglecting the interdependencies among retrieved documents. This oversight is particularly detrimental in multi-hop reasoning tasks requiring coherent logical relationships between documents~\cite{DBLP:journals/tacl/TrivediBKS22, DBLP:conf/coling/HoNSA20}.

In this work, we propose a novel end-to-end strategy for training rerankers in RAG systems. We reformulate the reranking task through the lens of attention masks, where selecting the top-$k$ subset from the retrieved candidate documents is viewed as the application of a document-wise top-$k$ attention mask during attention computation. This perspective leads to a shift in the problem formulation: instead of directly learning a more effective reranker, we focus on learning the optimal document-wise top-$k$ attention mask.

However, since the hard attention mask is discrete, it can not be directly optimized via gradient descent. To overcome this challenge, we introduce a solution based on the Gumbel Trick~\cite{DBLP:conf/iclr/JangGP17} and Relaxed Top-$k$ techniques~\cite{DBLP:conf/icml/ChenSWJ18}. This enables us to design a \textit{stochastic}, top-$k$ attention mask that is fully differentiable, allowing for end-to-end optimization. We note this approach as \textbf{D}ifferentiable \textbf{M}asked \textbf{A}ttention (DMA).
% , which makes the attention computation differentiable even with discrete document selection.

With DMA in place, we reformulate the reranking problem as learning the optimal sampling weight for the corresponding attention mask. This leads to our end-to-end training framework, which we refer to as \textbf{G}umbel \textbf{Rerank}ing (G-Rerank). Unlike previous methods that rely on LLM-supervised losses, G-Rerank directly optimizes the reranker by minimizing the overall language modeling loss of the RAG system, thereby ensuring that the training objective closely aligns with the inference process. Additionally, G-Rerank accounts for interdependencies between retrieved candidate documents, making it suitable for multi-hop QA tasks.

We evaluate our training approach across various architectures. Specifically, we conduct experiments using two language models—FiD~\cite{DBLP:conf/eacl/IzacardG21} and CEPE-Llama2-7B~\cite{DBLP:conf/acl/YenG024}—as well as two rerankers—BGE-Reranker-Base~\cite{xiao2023bge} and RankT5~\cite{DBLP:conf/sigir/Zhuang0J0MLNWB23}. Our method is tested on five benchmark datasets, covering both single-hop and multi-hop QA tasks. To comprehensively assess the effectiveness of our approach, we consider three different evaluation settings: \textit{mining}, \textit{reranking}, and \textit{generation}. Our proposed training strategy achieves consistent improvements across all these settings. Furthermore, compared to distillation-based methods, our training approach significantly improves the reranker's ability to distinguish \textit{indirectly relevant documents}, leading to a 10.4\% improvement in the Recall@5 metric on HotpotQA. Finally, we analyze the necessity of the Gumbel trick and the impact of prior knowledge in rerankers.

\section{Related Works}

\paragraph{Training the Reranking Module in RAG Systems}
The effectiveness of RAG systems relies heavily on the quality of retrieval and reranking ~\cite{DBLP:conf/naacl/GlassRCNCG22, DBLP:journals/corr/abs-2405-18414}. Traditional retrieval methods are based on lexical similarity ~\cite{DBLP:journals/ftir/RobertsonZ09}, while recent advances leverage dense vectors and transformer architectures~\cite{DBLP:conf/emnlp/KarpukhinOMLWEC20, DBLP:conf/sigir/KhattabZ20}. However, retrieval modules fine-tuned on public datasets often require additional adaptation for specific downstream tasks~\cite{DBLP:journals/jmlr/IzacardLLHPSDJRG23, DBLP:conf/sigir/SalemiZ24a}.

To bridge this gap, recent works explore fine-tuning retrieval and reranking modules for tasks such as open-domain question answering (ODQA). One common strategy distills knowledge from LLMs into retrievers by ranking candidate documents based on generated answer perplexity~\cite{DBLP:conf/naacl/ShiMYS0LZY24, DBLP:conf/naacl/GlassRCNCG22, DBLP:conf/iclr/Lin0CSL00KSLZY24}. However, such methods overlook inter-document dependencies, crucial for multi-hop reasoning tasks~\cite{DBLP:journals/tacl/TrivediBKS22, DBLP:conf/coling/HoNSA20}. Alternative approaches use attention scores~\cite{DBLP:conf/iclr/IzacardG21} or leave-one-out methods~\cite{DBLP:journals/jmlr/IzacardLLHPSDJRG23, DBLP:conf/naacl/Asai0H22}, but these are not end-to-end optimized for generation quality, leading to a retriever-generation gap~\cite{DBLP:conf/acl/KeK00MB24}.

\paragraph{Stochastic k-Subset Selection and Masked Attention}
Top-$k$ relaxation has been widely studied for differentiable subset sampling, extending the Gumbel-Softmax trick~\cite{DBLP:conf/iclr/JangGP17, DBLP:conf/ijcai/XieE19, DBLP:conf/nips/XieDCDZZ0P20}, with important applications in  semi-structured pruning~\cite{DBLP:conf/nips/FangYMHPK0W24}, model interpretability~\cite{DBLP:conf/icml/ChenSWJ18} and point clouds analysis~\cite{DBLP:conf/cvpr/YangZNLLZT19}.

The reranking process can also be modeled as a subset sampling problem. However, since retrieved documents influence LLM outputs through attention, a key challenge lies in introducing sparsity into the attention computation. Existing approaches employ soft attention masks to model discrete selections~\cite{DBLP:conf/naacl/FanGLWWJDZH21, DBLP:conf/cvpr/YangZNLLZT19}. Inspired by these methods, we model RAG reranking as a subset sampling process with soft masks, facilitating end-to-end optimization.

\begin{center}
\begin{algorithm*}
\caption{Gumbel Reranking: Training Reranker via Differentiable Masked Attention}
\label{alg:ranker_as_selector}
\begin{algorithmic}[1]
\Procedure{StochasticSubsetMask}{reranker $\mathcal{R}$, documents $\mathbf{d_1}, \dots, \mathbf{d_n}$, query $\mathbf{q}$, temperature $\tau$, scale factor $\kappa$, subset size $k$}
    \State $w_i = \mathcal{R}( \text{Concatenate}(\mathbf{q}, \mathbf{d}_i))\quad \forall i \in [n] \triangleq \{1, 2, \dots, n\}$  \Comment{Apply Reranker}

    \For{$j = 1$ \textbf{to} $k$} 
    \Comment{Stochastic Top-$k$ Sampling}
        \State $\tilde{w}_i = -\log(-\log(u_i)) + \kappa \cdot w_i,\quad u_i \sim \mathcal{U}(0, 1) \quad \forall i \in [n]$ 
        \State $\hat{\mathcal{M}}^{\mathcal{R},j} = \text{softmax}\left(\frac{\tilde{\mathbf{w}}}{\tau}\right), \quad \tilde{\mathbf{w}} = (\tilde{w}_1, \tilde{w}_2, \ldots, \tilde{w}_n)$
    \EndFor

    \State \Return $\max ( \hat{\mathcal{M}}^{\mathcal{R},1}, \dots, \hat{\mathcal{M}}^{\mathcal{R},k} )$ \Comment{Return Relaxed Top-$k$ Mask}
\EndProcedure
\State
\For{each $(\text{query } \mathbf{q}, \text{answer } \mathbf{a})$ in training data} \Comment{Training Loop}
    \State Retrieve $n$ documents $\mathbf{d_1}, \dots, \mathbf{d_n}$ using $\mathbf{q}$
    \State $\hat{\mathcal{M}}^\mathcal{R} = \operatorname{StochasticSubsetMask}(\mathcal{R}, \mathbf{d_1}, \dots, \mathbf{d_n}, \mathbf{q}, \tau, \kappa, k)$
    \State Apply $\mathcal{DMA}$($\hat{\mathcal{M}}^\mathcal{R}$) to obtain logits and language loss $\mathcal{L}_{LM}$  \Comment{\autoref{sec: DMA}}
    \State Update reranker $\mathcal{R}$ with $\nabla_{\mathcal{R}} \mathcal{L}_{LM}$ \Comment{Reranker Optimization}
\EndFor

\end{algorithmic}
\end{algorithm*}
\end{center}
\section{Methodology}

\subsection{Problem Setting}
For common downstream tasks, such as Open-Domain QA~\cite{DBLP:journals/corr/abs-2101-00774}, the training data typically consists of an input query $\mathbf{q}$ and the corresponding ground-truth answer $\mathbf{a}$. During the retrieval process, a set of candidate documents $\mathbf{d_1}, \dots, \mathbf{d_n}$ is retrieved based on $\mathbf{q}$. The reranker $\mathcal{R}$ is then applied to these candidate documents, generating a set of candidate scores. We retain only the top-\(k\) scored documents for further computation:
\begin{equation}
\begin{aligned}\label{eq: reranker forward pass}
    w_i &= \mathcal{R}(\text{Concatenate}(\mathbf{q}, \mathbf{d}_i)), \quad \forall i \in [n] \\
    \mathcal{I}_k &=\left\{ i \mid w_i \in \text{top-}k \left( \{ w_i \}_{i=1}^n \right) \right\}
\end{aligned}
\end{equation}
where $[n] \triangleq \{1, 2, \dots, n\}$. The top-$k$ documents, $\mathcal{D}_k= \{\mathbf{d}_i \mid i \in \mathcal{I}_k \}$, are selected as input to the LLM, which then computes the corresponding logits and language loss $\mathcal{L}_{LM}$. In this work, we focus on training the reranker in the RAG system. A key challenge is that the candidate documents $\mathbf{d_1}, \dots, \mathbf{d_n}$ lack relevance annotations, making it infeasible to directly fine-tune the reranker. Additionally, although we have access to the language loss $\mathcal{L}_{LM}$, the top-$k$ operation in \autoref{eq: reranker forward pass} is non-differentiable, preventing gradient propagation to the reranker and thus hindering end-to-end training.

\subsection{Viewing Reranker as Attention Mask}\label{sec: Viewing Reranker as Attention Mask}
We reinterpret the reranking process from the perspective of attention masks. Let $K_{i,t}$ and $V_{i,t}$ denote the key and value embeddings of the $t$-th token in the $i$-th candidate document, respectively. And let $Q_m$ denote the query embedding for the $m$-th token in the decoding phase of LLM, the standard attention computation is defined as:

\begin{equation}\label{eq: original attn}
    \mathcal{A}(Q_m, K_{i,t}) = \frac{\exp \left(\frac{Q_m K_{i,t}^T}{\sqrt{d_k}}\right)}{\sum_{i'} \sum_{t'} \exp \left(\frac{Q_m K_{i',t'}^T}{\sqrt{d_k}}\right)}
\end{equation}
where $\mathcal{A}(Q_m, K_{i,t})$ represents the attention score of the $m$-th token in the decoding process to the $t$-th token in the $i$-th candidate document.

The reranker retains only the top-\( k \) documents for attention computation, and these top-\( k \) documents, as a set, are used as part of the prompt. The order of these documents is no longer important. Therefore, we can use a corresponding \textit{hard} attention mask \( \mathcal{M}^{\mathcal{R}} \) to simulate the reranking process.

\begin{equation}\label{eq: hard attention mask}
    \mathcal{M}^{\mathcal{R}}_i = \begin{cases}
        1, & \text{if } i \in \mathcal{I}_k \\
        0, & \text{otherwise}
    \end{cases}
\end{equation}
% When LLM processes retrieved documents, those not selected by the reranker do not contribute to the attention computation. This process can be formally described using masked attention:

\begin{equation}
    \mathcal{MA}(Q_m, K_{i,t}) = \frac{\mathcal{M}^{\mathcal{R}}_i \exp \left(\frac{Q_m K_{i,t}^T}{\sqrt{d_k}}\right)}{\sum_{i'} \mathcal{M}^{\mathcal{R}}_{i'} \sum_{t'} \exp \left(\frac{Q_m K_{i',t'}^T}{\sqrt{d_k}}\right)}
\end{equation}

This formulation of masked attention is mathematically equivalent to reranking. If document $i$ is not selected by the reranker, i.e., $\mathcal{M}^{\mathcal{R}}_i = 0$, then all tokens within document $i$ receive an attention score of zero, i.e., $\mathcal{MA}(Q_m, K_{i,t}) = 0,~ \forall t$.

\paragraph{Independence Requirements in Pre-Filling} 
To effectively simulate reranking via $\mathcal{MA}$, it is crucial to ensure the independence of candidate documents. First, all candidate documents should use the same positional encoding to eliminate position bias. Second, each document should be encoded independently during pre-filling to prevent information leakage across documents. To enforce these constraints, we adopt the parallel pre-filling architecture, as seen in models like FiD~\cite{DBLP:conf/eacl/IzacardG21}, CEPE~\cite{DBLP:conf/acl/YenG024}, and Parallel Windows~\cite{DBLP:conf/acl/RatnerLBRMAKSLS23}, where retrieved documents are encoded separately with independent position encodings during the pre-filling stage.

% \paragraph{Independence Requirements in Pre-Filling} 
% Fortunately, we can employ the Gumbel trick to design a \textit{soft} attention mask, enabling differentiability and facilitating direct optimization through backpropagation.

% \paragraph{Stochastic Subset Mask}
\subsection{Differentiable Masked Attention}\label{sec: DMA}
The problem of learning a more effective reranker is thus reformulated as learning a better attention mask $\mathcal{M}^{\mathcal{R}}$. However, the \textit{hard} attention mask $\mathcal{M}^{\mathcal{R}}$ defined in \autoref{eq: hard attention mask} remains non-differentiable, preventing end-to-end optimization based on the final language loss. To solve this problem, we leverage the Gumbel-Softmax technique~\cite{DBLP:conf/iclr/JangGP17} to convert discrete sampling into a differentiable process. Specifically, we transform the reranker’s output $w_1, w_2, \dots, w_n$ into a probability distribution for sampling an attention mask:
\begin{equation}
\begin{aligned}\label{eq: single sample}
  G_i &= -\log\Bigl(-\log(u_i)\Bigr), \quad u_i \sim \mathcal{U}(0,1),\\
  \hat{\mathcal{M}}^\mathcal{R} &= \frac{\exp\!\Bigl(\tfrac{\tilde{w}_i}{\tau}\Bigr)}
              {\sum_{j=1}^{n}\exp\!\Bigl(\tfrac{\tilde{w}_j}{\tau}\Bigr)}, \quad \tilde{w}_i = G_i + \kappa \cdot w_i
\end{aligned}
\end{equation}
where $\hat{\mathcal{M}}^\mathcal{R}_i$ represents the probability of selecting the $i$-th document. \(\tau\) and \(\kappa\) are hyperparameters in the Gumbel training process. We discuss their effects in detail in \autoref{sec: Effect of hyper-parameters on the Training Process}. To approximate Top-$k$ reranking, we perform independent sampling $k$ times and compute the element-wise maximum:

\begin{equation}
    \hat{\mathcal{M}}^{\mathcal{R}} = \max \left( \hat{\mathcal{M}}^{\mathcal{R},1}, \dots, \hat{\mathcal{M}}^{\mathcal{R},k} \right)
    \label{eq:subset_sampling}
\end{equation}

This results in a \textit{soft} attention mask representing the sampled subset, leading to Differentiable Masked Attention:

\begin{equation}\label{eq: DMA}
    \mathcal{DMA}(Q_m, K_{i,t}) = \frac{\hat{\mathcal{M}}^{\mathcal{R}}_i \exp \left(\frac{Q_m K_{i,t}^T}{\sqrt{d_k}}\right)}{\sum_{i'} \hat{\mathcal{M}}^{\mathcal{R}}_{i'} \sum_{t'} \exp \left(\frac{Q_m K_{i',t'}^T}{\sqrt{d_k}}\right)}
\end{equation}

This formulation allows end-to-end optimization of the reranker $\mathcal{R}$ based on final language model loss, improving overall RAG system performance.

\begin{figure*}[!th]
  \includegraphics[width=0.99\linewidth]{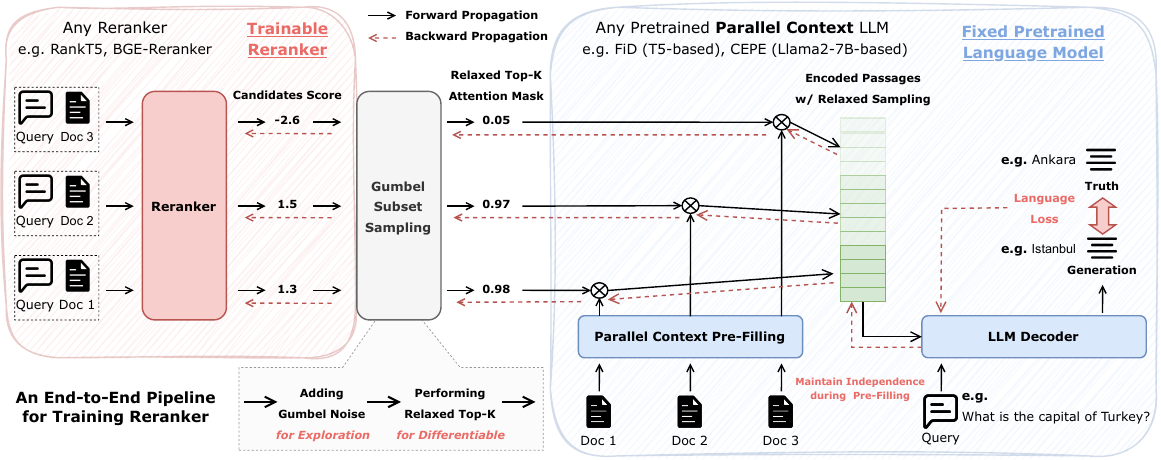}
  \centering
\caption{Implementation workflow of G-Rerank. We focus on fine-tuning the reranker while keeping LLM parameters fixed. However, given sufficient computational resources, joint fine-tuning of both the LLM and the reranker is feasible. In the Pre-Filling phase, it is essential to maintain the independence of candidate documents.}
\label{fig:method}
\end{figure*}

\subsection{Gumbel Reranking Pipeline}

In this section, we introduce \textit{Gumbel Reranking}, an end-to-end reranker optimization framework leveraging the previously introduced $\mathcal{DMA}$. The overall pipeline is outlined in \hyperref[alg:ranker_as_selector]{Algorithm~\ref{alg:ranker_as_selector}}.

\paragraph{Training Process} 
Given a query $\mathbf{q}$ and a set of candidate documents $\mathbf{d_1}, \dots, \mathbf{d_n}$, the reranker first computes a relevance score for each document. The Stochastic Subset Mask algorithm then generates a Top-$k$ attention mask $\hat{\mathcal{M}}^{\mathcal{R},k}$, which represents the probability of selecting each candidate document. The selected documents are subsequently used in the generation process, where the attention mechanism follows \autoref{eq: DMA} to compute logits and the language modeling loss. Finally, the reranker is optimized by minimizing the language loss $\mathcal{L}_{LM}$. Since our proposed framework primarily focuses on enhancing the reranking module, we fix the parameters of the LLM in our experimental setup, as shown in \autoref{fig:method}. This also facilitates a fairer comparison of different reranker training strategies.

\paragraph{Key Advantages} 
This framework facilitates end-to-end optimization of the reranker via backpropagation, offering two primary advantages. First, by modeling reranking as applying a document-wise attention mask, it mitigates the discrepancy between training and inference, guiding the reranker to prioritize documents that minimize the final generation loss. Second, our approach leverages gumbel subset sampling, enabling the model to identify the complete evidence \textit{subset} during training, rather than analyzing each candidate document independently. This advantage makes our method well-suited for multi-hop QA scenarios and sets it apart from existing perplexity-based distillation techniques, as discussed in \autoref{sec: Challenges in Handling Indirectly Relevant Documents with EMDR/PDist}.

\section{Experiments}

\begin{figure*}[t]
\centering
  \includegraphics[width=\linewidth]{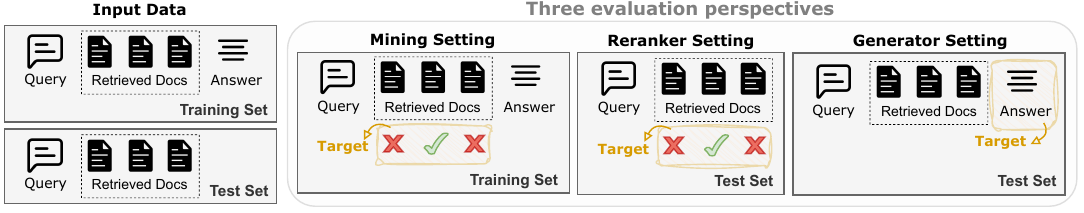}
  \caption{Comparison of three different experimental settings. In addition to common evaluation metrics on the test set, we also assess the reranker's ability to identify relevant documents from the training set.}
  \label{fig:expsetting}
\end{figure*}

\begin{table*}[!ht]
\centering
\small
\resizebox{0.8\linewidth}{!}{
\begin{tabular}{ l || c c|| c c | c || c c c }
    \toprule
    & \multicolumn{2}{c||}{\textbf{Mining Setting}} & \multicolumn{3}{c||}{\textbf{Reranker Setting}} & \multicolumn{3}{c}{\textbf{Generator Setting}} \\  
    \cmidrule(r){2-3} \cmidrule(r){4-6} \cmidrule(r){7-9}
    \bf Training Methods & \textbf{Recall@5} & \textbf{NDCG@5} & \textbf{Recall@5} & \textbf{NDCG@5} & \textbf{MRR} & \textbf{EM} & \textbf{SubEM} & \textbf{F1} \\
    \addlinespace[0.2em]
    \hline
    \hline
    \addlinespace[0.4em]
    \multicolumn{9}{c}{\textbf{Dataset: }\normalsize\texttt{Hotpotqa}} \\[-0.1em]
    \addlinespace[0.1em]\multicolumn{1}{l}{\bf{Reranker: }\normalsize\texttt{RankT5}} & \multicolumn{8}{l}{}\\[0.2em]
    - EMDR~\cite{DBLP:conf/iclr/Lin0CSL00KSLZY24} & \underline{78.0} & \underline{80.5} & \underline{78.7} & \underline{80.6} & \bf 95.9 & \underline{60.8} & \underline{66.1} & \underline{75.8} \\
    - PDist~\cite{DBLP:conf/naacl/GlassRCNCG22} & 76.8 & 79.5 & 78.1 & 80.0 & \underline{95.7} & \underline{60.8} & 66.0 & \underline{75.8} \\
    - LOOP~\cite{DBLP:journals/jmlr/IzacardLLHPSDJRG23} & 71.7 & 74.7 & 72.5 & 74.9 & 93.0 & 60.0 & 65.1 & 75.0 \\
    - ADist~\cite{DBLP:conf/iclr/IzacardG21} & 71.3 & 72.1 & 71.3 & 71.9 & 88.4 & 57.0 & 61.9 & 71.5 \\
    \rowcolor{gray!10} - G-Rerank & \bf 83.3 & \bf 84.7 & \bf 84.4 & \bf 84.9 & \bf 95.9 & \bf 61.1 & \bf 66.5 & \bf 76.3 \\
    % \midrule
    \addlinespace[0.4em]
    % \multicolumn{9}{c}{\textbf{Dataset: }\normalsize\texttt{Hotpotqa}} \\[-0.1em]
    \addlinespace[0.1em]\multicolumn{1}{l}{\bf{Reranker: }\normalsize\texttt{BGE-Base}} & \multicolumn{8}{l}{}\\[0.2em]
    - EMDR~\cite{DBLP:conf/iclr/Lin0CSL00KSLZY24} & \underline{81.1} & \underline{83.2} &\bf{81.8} & \bf 83.1 & \bf{96.3} & \underline{60.8} & \underline{66.0} & \bf 75.8 \\
    - PDist~\cite{DBLP:conf/naacl/GlassRCNCG22} & 79.1 & 81.6 & 81.2 & 82.6 & \underline{96.2} & \bf 60.9 & \bf 66.1 & \underline{75.7} \\
    - LOOP~\cite{DBLP:journals/jmlr/IzacardLLHPSDJRG23} & 79.1 & 81.1 & 80.4 & 81.7 & 95.3 & 60.3 & 65.4 & 75.2 \\
    - ADist~\cite{DBLP:conf/iclr/IzacardG21} & 77.7 & 79.5 & 78.1 & 79.5 & 93.7 & 59.8 & 65.0 & 74.7 \\
    \rowcolor{gray!10} - G-Rerank & \bf 81.6 & \bf 83.3 & \underline{81.1} & \underline{82.9} & 95.8 & \bf 60.9 & \bf 66.1 & \underline{75.7} \\
    \addlinespace[0.4em]
    \hline
    \hline
    \addlinespace[0.4em]
    \multicolumn{9}{c}{\textbf{Dataset: }\normalsize\texttt{Musique}} \\[-0.1em]
    \addlinespace[0.1em]\multicolumn{1}{l}{\bf{Reranker: }\normalsize\texttt{RankT5}} & \multicolumn{8}{l}{}\\[0.2em]
    - EMDR~\cite{DBLP:conf/iclr/Lin0CSL00KSLZY24} & 56.6 & \underline{65.8} & \underline{55.0} & \underline{58.1} & \bf 82.0 & \underline{39.6} & 42.1 & \underline{48.6} \\
    - PDist~\cite{DBLP:conf/naacl/GlassRCNCG22} & \underline{57.3} & 65.3 & 52.7 & 55.0 & 79.5 & \underline{39.6} & \underline{42.2} & 48.3 \\
    - LOOP~\cite{DBLP:journals/jmlr/IzacardLLHPSDJRG23} & 56.3 & 64.9 & 53.3 & 55.6 & 79.6 & 39.2 & 41.7 & 48.0 \\
    - ADist~\cite{DBLP:conf/iclr/IzacardG21} & 53.8 & 55.3 & 47.7 & 47.3 & 66.4 & 35.4 & 37.9 & 44.1 \\
    \rowcolor{gray!10} - G-Rerank & \bf 60.7 & \bf 67.8 & \bf 57.9 & \bf 59.7 & \underline{81.5} & \bf 40.0 & \bf 42.4 & \bf 49.1 \\
    % \midrule
    \addlinespace[0.4em]
    % \multicolumn{9}{c}{\textbf{Dataset: }\normalsize\texttt{Musique}} \\[-0.1em]
    \addlinespace[0.1em]\multicolumn{1}{l}{\bf{Reranker: }\normalsize\texttt{BGE-Base}} & \multicolumn{8}{l}{}\\[0.2em]
    - EMDR~\cite{DBLP:conf/iclr/Lin0CSL00KSLZY24} & 56.6 & 65.7 & 53.6 & 57.1 & \underline{81.5} & \underline{39.7} & \underline{42.4} & \underline{48.8} \\ 
    - PDist~\cite{DBLP:conf/naacl/GlassRCNCG22} & \underline{60.3} & \underline{66.1} & \bf 58.2 & \underline{59.6} & 80.5 & 39.4 & 42.3 & 48.6 \\
    - LOOP~\cite{DBLP:journals/jmlr/IzacardLLHPSDJRG23} & 58.7 & 65.6 & 57.2 & 59.3 & \bf 81.8 & \underline{39.7} & 42.2 & \underline{48.8} \\
    - ADist~\cite{DBLP:conf/iclr/IzacardG21} & 57.9 & 64.5 & 46.0 & 45.3 & 64.7 & 34.8 & 37.3 & 43.4 \\
    \rowcolor{gray!10} - G-Rerank & \bf 60.9 & \bf 66.6 & \underline{57.6} & \bf 59.7 & \underline{81.5} & \bf 39.9 & \bf 42.7 & \bf 49.1 \\
    \addlinespace[0.4em]
    \hline
    \hline
    \addlinespace[0.4em]
    \multicolumn{9}{c}{\textbf{Dataset: }\normalsize\texttt{2wikihop}} \\[-0.1em]
    \addlinespace[0.1em]\multicolumn{1}{l}{\bf{Reranker: }\normalsize\texttt{RankT5}} & \multicolumn{8}{l}{}\\[0.2em]
    - EMDR~\cite{DBLP:conf/iclr/Lin0CSL00KSLZY24} & 58.6 & 63.4 & 62.9 & 68.7 & 88.7 & 67.2 & 69.9 & 72.5 \\
    - PDist~\cite{DBLP:conf/naacl/GlassRCNCG22} & 72.6 & 76.5 & 77.2 & 81.9 & 94.1 & 70.2 & 73.0 & 75.5 \\
    - LOOP~\cite{DBLP:journals/jmlr/IzacardLLHPSDJRG23} & \underline{80.4} & \bf 87.1 & \underline{79.2} & \underline{85.4} & \underline{97.5} & \underline{71.6} & \underline{74.4} & \underline{76.9} \\
    - ADist~\cite{DBLP:conf/iclr/IzacardG21} & 74.7 & 79.2 & 72.4 & 76.6 & 90.1 & 64.1 & 66.5 & 69.6 \\
    \rowcolor{gray!10} - G-Rerank & \bf 80.8 & \underline{86.9} & \bf 82.7 & \bf 88.4 & \bf 97.8 & \bf 71.8 & \bf 74.7 & \bf 77.2 \\
    % \midrule
    \addlinespace[0.4em]
    % \multicolumn{9}{c}{\textbf{Dataset: }\normalsize\texttt{2wikihop}} \\[-0.1em]
    \addlinespace[0.1em]\multicolumn{1}{l}{\bf{Reranker: }\normalsize\texttt{BGE-Base}} & \multicolumn{8}{l}{}\\[0.2em]
    - EMDR~\cite{DBLP:conf/iclr/Lin0CSL00KSLZY24} & 61.8 & 67.3 & 71.0 & 77.1 & 93.8 & 68.9 & 71.8 & 74.3 \\
    - PDist~\cite{DBLP:conf/naacl/GlassRCNCG22} & 74.0 & 76.8 & 76.6 & 82.2 & 94.5 & 69.1 & 71.9 & 74.4 \\
    - LOOP~\cite{DBLP:journals/jmlr/IzacardLLHPSDJRG23} & 77.3 & 85.0 & 76.0 & 83.3 & \bf 98.5 & \bf 71.2 & \bf 73.9 & \bf 76.3 \\
    - ADist~\cite{DBLP:conf/iclr/IzacardG21} & \bf 81.4 & \bf 87.7 & \underline{80.5} & \underline{86.4} & 97.1 & 70.7 & 73.5 & 76.1 \\
    \rowcolor{gray!10} - G-Rerank & \underline{79.6} & \underline{86.4} & \bf 81.4 & \bf 86.5 & \underline{97.5} & \underline{70.9} & \underline{73.7} & \underline{76.2} \\
    \bottomrule
\end{tabular}
}
\caption{Experiments on 2WikiHop, Musique, and HotpotQA using FiD-Large as reader. We consider the settings illustrated in \autoref{fig:expsetting}. The best performance is highlighted in bold, while the second-best performance is underlined.}
\label{tbl:multi dataset experiment}
\end{table*}

In \autoref{sec: main exp}, we first validate the effectiveness of our approach under three different experimental settings. Then, in \autoref{sec: indirect exp}, we focus on whether the reranker can learn to prioritize \textit{indirect evidence} in multi-hop question answering. 
% These are documents that belong to the evidence chain but do not directly contain the final correct answer. Our results demonstrate that our method is significantly more effective at capturing the importance of such indirect evidence. 
Next, in \autoref{sec: gumbel exp}, we conduct an ablation study on the Gumbel trick and demonstrate its necessity. 
% Without the Gumbel trick, the model tends to learn a ``uniform'' soft attention mask. 
Finally, in \autoref{sec: learnable weight exp}, we remove the reranker and assign each document a learnable weight to further verify the efficacy of our training objective in capturing the relative importance of documents.

\subsection{Experimental Setup}

\paragraph{Language Models}
We experiment with two different language models as the generation module in our RAG system: Fusion-in-Decoder (FiD)~\cite{DBLP:conf/eacl/IzacardG21} and CEPE-Llama2-7B~\cite{DBLP:conf/acl/YenG024}.  FiD~\cite{DBLP:journals/jmlr/RaffelSRLNMZLL20}, built upon the T5 architecture, is specifically designed for knowledge-intensive QA and is fine-tuned for each task. CEPE-Llama2-7B segments long documents with a lightweight encoder and employs cross-attention for effective context utilization, operating in a zero-shot manner.

\paragraph{Reranker}
We experiment with RankT5-Base~\cite{DBLP:conf/sigir/Zhuang0J0MLNWB23} and BGE-Base-Reranker~\cite{xiao2023bge} as the reranking module in the RAG system. RankT5-Base is fine-tuned in an encoder-decoder setup to perform reranking, while BGE-Base-Reranker is an encoder-only model based on BERT.

\paragraph{Datasets}
We evaluate on five QA datasets: multi-hop (2WikiHop~\cite{DBLP:conf/coling/HoNSA20}, HotpotQA~\cite{DBLP:conf/emnlp/Yang0ZBCSM18}, Musique~\cite{DBLP:journals/tacl/TrivediBKS22}) and single-hop (NQ~\cite{DBLP:journals/tacl/KwiatkowskiPRCP19}, TQA~\cite{DBLP:conf/acl/KimKK19}). Details are in \autoref{sec: appendix_datasets}. For NQ and TQA, we retrieve 20 candidate documents per query using DPR~\cite{DBLP:conf/emnlp/KarpukhinOMLWEC20}. For multi-hop datasets, we apply the distraction setting to ensure ground-truth documents are included, adding 10 random candidates in Musique to maintain 20 candidates per query.

\paragraph{Baselines}
We compare against four LLM-supervised reranker training methods: EMDR~\cite{DBLP:conf/nips/SachanRHDY21, DBLP:conf/naacl/ShiMYS0LZY24, DBLP:conf/iclr/Lin0CSL00KSLZY24}, PDist~\cite{DBLP:journals/jmlr/IzacardLLHPSDJRG23, DBLP:conf/naacl/GlassRCNCG22}, LOOP~\cite{DBLP:journals/jmlr/IzacardLLHPSDJRG23}, and ADist~\cite{DBLP:conf/iclr/IzacardG21}, which employ different LLM-supervised losses. Details about these baselines can be found in \autoref{sec: appendix_baseline}.
\begin{table*}[!th]
\centering
\small
\resizebox{\linewidth}{!}{
\begin{tabular}{l | c c c | c c c | c c c | c c c}
    \toprule
    & \multicolumn{6}{c|}{\textbf{Reranker Setting}} & \multicolumn{6}{c}{\textbf{Mining Setting}} \\
    \cmidrule(lr){2-7} \cmidrule(lr){8-13}
    & \multicolumn{3}{c|}{\textbf{FiD-Base}} & \multicolumn{3}{c|}{\textbf{FiD-Large}} & \multicolumn{3}{c|}{\textbf{FiD-Base}} & \multicolumn{3}{c}{\textbf{FiD-Large}} \\
    \cmidrule(lr){2-4} \cmidrule(lr){5-7} \cmidrule(lr){8-10} \cmidrule(lr){11-13}
    \textbf{Training Methods} & Recall & MRR & NDCG & Recall & MRR & NDCG & Recall & MRR & NDCG & Recall & MRR & NDCG \\
    \midrule
    - EMDR~\cite{DBLP:conf/iclr/Lin0CSL00KSLZY24}  & \underline{63.0} & \underline{45.6} & \underline{45.2} & \underline{61.8} & \underline{45.2} & \underline{44.4} & \underline{60.3} & \underline{42.9} & \underline{42.3} & \underline{59.0} & \underline{42.7} & \underline{41.6} \\
    - PDist~\cite{DBLP:conf/naacl/GlassRCNCG22}  & 50.5 & 39.8 & 36.2 & 60.2 & 44.4 & 43.4 & 47.6 & 37.5 & 33.5 & 56.3 & 41.3 & 39.7 \\
    - LOOP~\cite{DBLP:journals/jmlr/IzacardLLHPSDJRG23}  & 53.1 & 40.7 & 38.1 & 52.5 & 40.2 & 37.3 & 49.6 & 38.2 & 34.9 & 49.8 & 37.6 & 34.5 \\
    - ADist~\cite{DBLP:conf/iclr/IzacardG21}  & 55.2 & 43.4 & 40.8 & 56.3 & 44.5 & 41.9 & 52.8 & 41.5 & 38.5 & 54.2 & 42.3 & 39.5 \\
    \rowcolor{gray!10} - G-Rerank  & \bf 69.3 & \bf 48.2 & \bf 49.6 & \bf 72.2 & \bf 49.5 & \bf 51.5 & \bf 65.5 & \bf 45.0 & \bf 45.8 & \bf 68.4 & \bf 46.4 & \bf 47.8 \\
    \bottomrule
\end{tabular}
}
\caption{Results on HotpotQA using FiD as reader for identifying \textit{indirectly relevant documents}, which are part of the evidence chain but do not directly contain the answer. Details can be found in \autoref{sec: appendix_indirectly relevant documents setting}.}

\label{tbl:hotpotqa-indirect-combined}
\end{table*}

\begin{table}[!ht]
\centering
\small
\resizebox{0.95\linewidth}{!}{
\begin{tabular}{l | c c | c c}
    \toprule
    & \multicolumn{2}{c|}{\textbf{RankT5}} & \multicolumn{2}{c}{\textbf{BGE-Base}} \\
    \cmidrule(lr){2-3} \cmidrule(lr){4-5}
    \textbf{Training Methods} & NQ & TQA & NQ & TQA \\
    \midrule
    - EMDR~\cite{DBLP:conf/iclr/Lin0CSL00KSLZY24}  & 33.4 & \underline{62.4} & 33.7 & \underline{62.5} \\
    - PDist~\cite{DBLP:conf/naacl/GlassRCNCG22}  & 32.9 & 61.8 & \underline{33.9} & 61.7 \\
    - LOOP~\cite{DBLP:journals/jmlr/IzacardLLHPSDJRG23}  & \underline{33.7} & 62.1 & 33.5 & 62.2 \\
    - ADist~\cite{DBLP:conf/iclr/IzacardG21}  & 33.1 & 61.6 & 33.2 & 62.0 \\
    \rowcolor{gray!10} - G-Rerank  & \bf 34.3 & \bf 62.8 & \bf 34.5 & \bf 63.1 \\
    \bottomrule
\end{tabular}
}
\caption{Experimental results on NQ and TQA datasets using CEPE-Llama2-7B as the reader. We employ SubEM as the evaluation metric.}
\label{tbl:cepe-nq-tqa-subem}
\end{table}

\subsection{Main Experiments}\label{sec: main exp}

\paragraph{Task Definition}
We consider the QA task where the model is trained on question-answer pairs along with retrieved documents, but at test time, it only receives the question and the retrieved documents. We define three evaluation settings, with their respective distinctions illustrated in \autoref{fig:expsetting}:

\begin{enumerate}
    \item \textbf{Mining Setting}: During training, given a \textit{question-answer pair}, can the reranker effectively identify relevant documents?
    \item \textbf{Reranker Setting}: At test time, given a \textit{question}, can the reranker effectively identify relevant documents?
    \item \textbf{Generator Setting}: At test time, given a \textit{question}, can the model generate correct answers?
\end{enumerate}

% \autoref{fig:expsetting} illustrates the three experimental settings. Since our primary focus in this work is on the training strategy for the reranker, we keep the parameters of the generation module fixed and train only the reranker in our comparative experiments.

\paragraph{Experimental Results}
\autoref{tbl:multi dataset experiment} presents the experimental results using FiD-Large as the generator model. Our method, G-Rerank, achieves the best or second-best performance across most datasets. In the Mining Setting, G-Rerank significantly improves the ability to identify relevant documents during training, given question-answer pairs. For instance, it achieves a 5.3\% improvement on the \texttt{HotpotQA} when using RankT5. In the Reranker Setting, G-Rerank demonstrates a notable improvement over other LLM-supervised loss-based training methods, with a 5.7\% Recall improvement on \texttt{HotpotQA} when using RankT5. Furthermore, in the Generator Setting, G-Rerank shows consistent performance gains in generation quality, as G-Rerank directly takes the  minimization of the final generation loss as the training objective.

\autoref{tbl:cepe-nq-tqa-subem} presents the SubEM results using CEPE-Llama2-7B as the generator model. We do not fine-tune CEPE-Llama2-7B on the downstream datasets; instead, we leverage its zero-shot capabilities. On both \texttt{NQ} and \texttt{TQA}, the G-Rerank training strategy leads to the best generation performance. Notably, these improvements are achieved solely by fine-tuning the retrieval module while keeping the language model parameters fixed.

% By performing end-to-end fine-tuning of the reranker with the objective of minimizing the final language loss in answer generation, G-Rerank not only improves the reranker's ability to distinguish relevant documents but also enhances the quality of the generator. This results in an overall performance boost in the RAG-based QA system.

\begin{figure*}[!ht]
\centering
\begin{minipage}{0.4\linewidth}
  \centering
  \includegraphics[width=\linewidth]{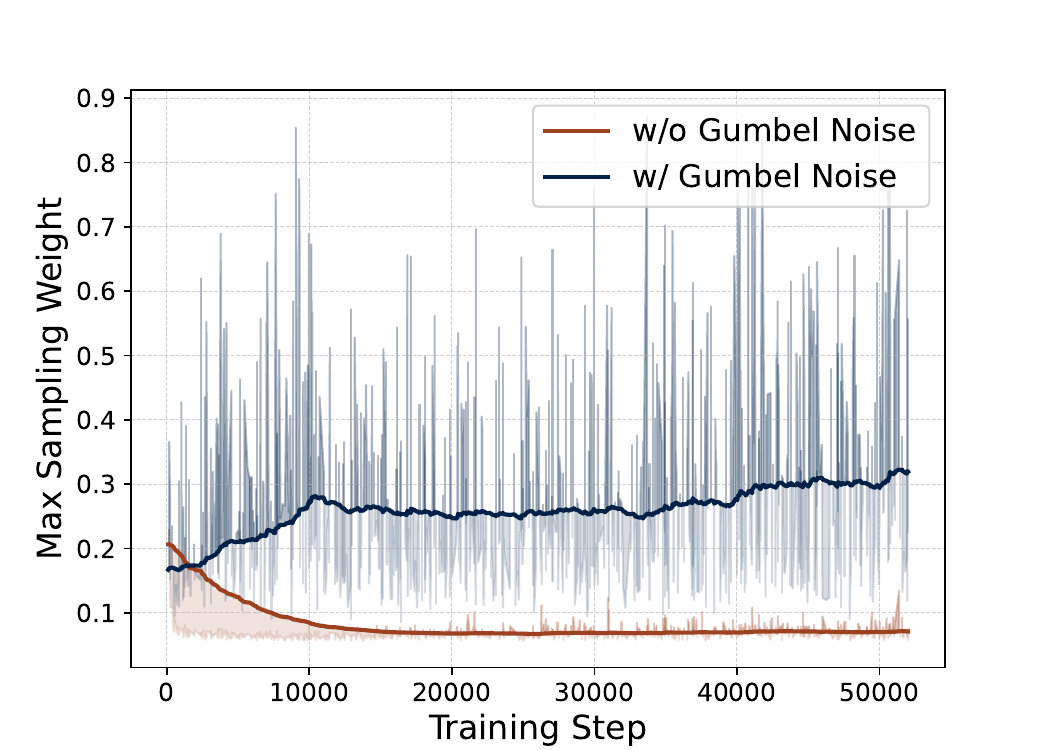}
  \caption{Comparison of Max Sampling Weight (indicating the reranker's ability to distinguish between candidate documents) with and without Gumbel noise on the NQ dataset.}
  \label{fig:gumbel_masking}
\end{minipage}
\begin{minipage}{0.04\linewidth}
\end{minipage}
\begin{minipage}{0.55\linewidth}
  \centering
  \includegraphics[width=\linewidth]{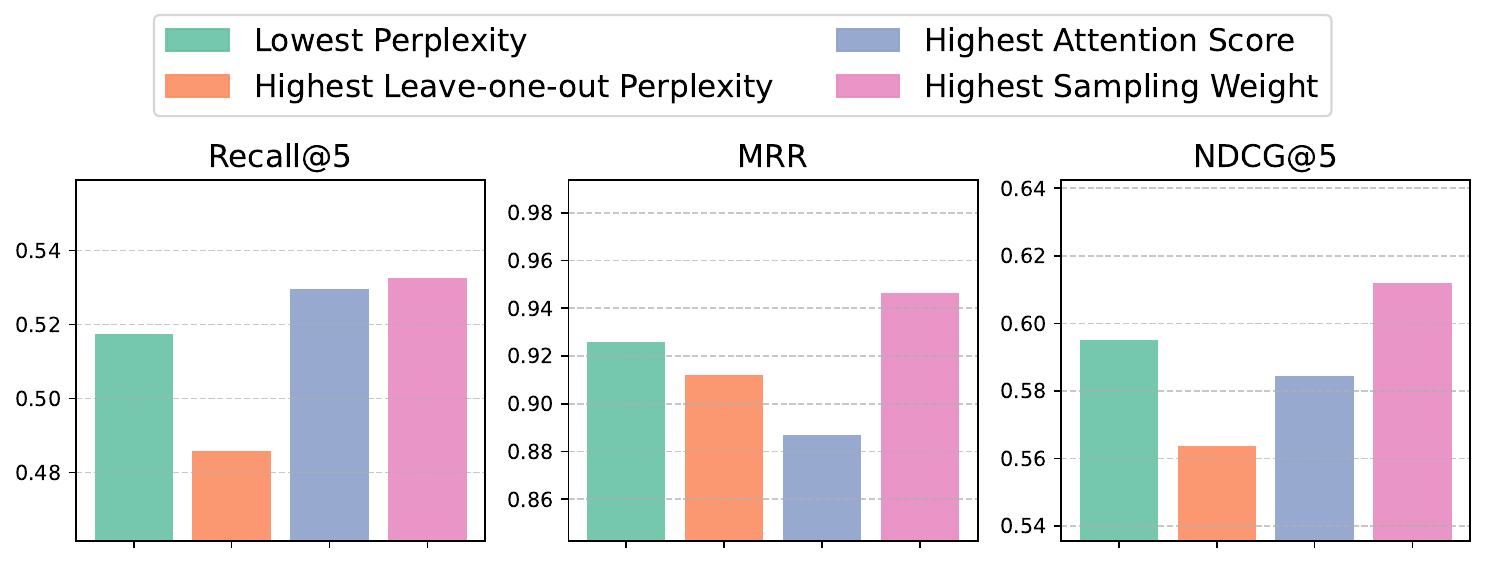}
  \caption{Performance comparison of different scalar metrics for assessing candidate document relevance in the Mining Setting. Our method is illustrated in \autoref{fig:learnable_sampling_weight method} and \hyperref[alg:masking_no_reranker]{Algorithm~\ref{alg:masking_no_reranker}}, while other baseline methods are described in detail in \autoref{sec: appendix_Learnable Sampling Weights}.}
  \label{fig:Performance_Comparison_DOC_Metrics}
\end{minipage}%
\end{figure*}

\subsection{Identifying Indirectly Relevant Documents}\label{sec: indirect exp}

In multi-hop question answering, a RAG system is required to retrieve a complete evidence chain comprising multiple documents to support its final answer. In such scenarios, the reranker should be able to identify \textit{indirectly relevant documents}, which are relevant to the query but do not directly contain the final answer. The challenge, however, lies in the fact that these documents often serve as `partial' evidence, and their relevance is not immediately apparent without being combined with other documents. Existing perplexity-based training methods commonly used in the literature distill independent relevance scores for each document, which fail to capture the inter-document dependencies that are essential for identifying indirectly relevant documents, as discussed in \autoref{sec: Challenges in Handling Indirectly Relevant Documents with EMDR/PDist}.

We evaluate various reranker training methods on \texttt{HotpotQA} to assess their ability to identify \textit{indirectly relevant documents}. To obtain the data, we employ a straightforward rule-based method to extract such documents: any document labeled as relevant in the dataset but not directly containing the final answer is considered an indirectly relevant document. Further discussion about this rule can be found in \autoref{sec: appendix_indirectly relevant documents setting}.

The experimental results are summarized in \autoref{tbl:hotpotqa-indirect-combined}. Our method, G-Rerank, demonstrates a significant improvement in identifying indirectly relevant documents. Specifically, when FiD-Large is used as the generator model, G-Rerank achieves a recall improvement of 10.4\%. These results suggest that our approach, which views reranking as a subset sampling problem, allows the model to better capture inter-document relationships and effectively recognize complete evidence chains.

\subsection{Necessity of Gumbel Trick}\label{sec: gumbel exp}

We leverage the Gumbel trick to transform the output weights of the reranker into an approximately discrete attention mask, where values tend to converge to either 0 or 1. A natural question arises: \textit{Is the introduction of Gumbel noise essential?} We conduct an ablation study by removing the Gumbel noise and directly utilizing the reranker's output weights as the attention mask while maintaining the same end-to-end optimization process.

Our experiments reveal a substantial drop in performance when Gumbel noise is omitted. Specifically, the EM metric on the \texttt{NQ} dataset decreases drastically from \( 46.2 \) (with Gumbel) to \( 12.7 \) (without Gumbel). To gain further insight, we visualize the reranker's output weights during training.

\autoref{fig:gumbel_masking} presents the average maximum normalized document weight assigned by the reranker. With the Gumbel noise applied, we observe a clear upward trend in the maximum document weight, indicating that the reranker progressively enhances the differentiation between candidate documents, which ultimately leads to convergence. In contrast, when Gumbel noise is removed, the maximum document weight decreases over time, eventually stabilizing at \( 0.05 \), signaling a diminished ability to distinguish between candidates. This degradation occurs because, in the absence of the discretization constraint introduced by the Gumbel trick, the model tends to preserve the original attention distribution, thus treating the removal of the attention mask as its objective. Consequently, the reranker learns to assign uniform soft mask across all candidates, i.e., \( \hat{\mathcal{M}}^{\mathcal{R}}_{i, \text{w/o Gumbel}}=\frac{1}{N}=0.05, \forall i \), thereby reverting the masked attention mechanism \autoref{eq: DMA} to its original form as defined in \autoref{eq: original attn}. These findings underscore the importance of the discretization constraint imposed by the Gumbel trick for learning an effective attention mask.

% \autoref{fig:gumbel_masking} illustrates the maximum normalized document weight assigned by the reranker. When Gumbel noise is applied, we observe a clear upward trend in the maximum document weight, indicating that the reranker progressively enhances the distinction between candidate documents, ultimately leading to convergence. This suggests that the model is continuously improving the probability of sampling the most relevant document. In contrast, without Gumbel noise, the maximum document weight decreases over time, converging towards \( 0.05 \), which indicates a reduced ability to distinguish between candidate documents. This degradation occurs because, in the absence of the discretization constraint introduced by the Gumbel trick, the model tends to preserve the original attention distribution, effectively treating the removal of the attention mask as its objective. Consequently, the model learns to assign uniform attention weights across all candidates, i.e., \( \hat{\mathcal{M}}^{\mathcal{R}}_{i, \text{w/o Gumbel}}=\frac{1}{N}=0.05, \forall i \), thereby reverting \autoref{eq: DMA} to \autoref{eq: original attn}. These results demonstrate that the discretization constraint introduced by the Gumbel trick is crucial for learning an effective attention mask.

\subsection{Learnable Sampling Weights}\label{sec: learnable weight exp}
The presence of the reranker can be viewed as incorporating text-based prior knowledge into the document relevance learning process. However, even in the absence of text-based priors, our training methodology can still effectively identify the relevant documents. To verify this, we focus on the Mining Setting and investigate whether the model is capable of learning meaningful document relevance scores \textit{without the use of a reranker}.

Our method is illustrated in \autoref{fig:learnable_sampling_weight method} and \hyperref[alg:masking_no_reranker]{Algorithm~\ref{alg:masking_no_reranker}}, while the experimental setup and baseline methods are explained in \autoref{sec: appendix_Learnable Sampling Weights}. Specifically, we remove the reranker component and instead assign each candidate document a learnable sampling weight, initializing all weights to zero. The results, presented in \autoref{fig:Performance_Comparison_DOC_Metrics}, show that even without the reranker (i.e., without prior knowledge of the text), our approach is still able to learn reliable relevance scores for each document. Moreover, it significantly outperforms other scalar metrics based on perplexity or attention scores, further confirming the effectiveness of our training objective.

\section{Conclusion}
In this work, we introduce G-Rerank, an end-to-end optimization framework for training rerankers in RAG systems. By reinterpreting the reranking process as masked attention, we leverage the Gumbel Trick and Relaxed Top-\( k \) to enable direct optimization of the document-wise attention mask. Our method effectively captures document interdependencies and aligns retrieval and generation objectives. Experiments across different settings show that G-Rerank notably improves reranker performance, especially in multi-hop QA tasks.
\section{Limitations}

Our method imposes certain constraints on its applicability to existing decoder-only LLMs due to its reliance on parallel encoding/decoding capabilities during the pre-filling stage. This requirement limits its direct adoption in conventional autoregressive LLMs. However, it is worth noting that many high-performance language models with parallel encoding/decoding capabilities have already become standard choices in various Retrieval-Augmented Generation (RAG) systems, such as FiD~\cite{DBLP:conf/eacl/IzacardG21}, CEPE~\cite{DBLP:conf/acl/YenG024}, and Parallel Windows~\cite{DBLP:conf/acl/RatnerLBRMAKSLS23}. Furthermore, our approach requires such models only during the reranker training phase; once trained, the reranker itself is independent of any specific LLM and can be flexibly adapted to other decoder-only models. Therefore, our method primarily serves as a general training framework rather than imposing architectural constraints on the final inference model. Additionally, our approach introduces extra hyperparameters in the Gumbel-Softmax process, including the temperature parameter $\tau$ and the scaling factor $\kappa$, which require tuning to achieve optimal performance. However, through empirical studies, we find that $\tau=0.5$ and $\kappa=1.0$ provide robust and stable performance across different model architectures and datasets. We provide a further discussion on the effect of $\tau$ and $\kappa$ in \autoref{sec: Effect of hyper-parameters on the Training Process}.

\section{Ethical Considerations}
While our method aims to improve the accuracy of the RAG system, it does not eliminate the inherent risks of biased data or model outputs, as the performance of RAG systems still heavily depends on the quality of training data and underlying models. The potential for bias in the training data, particularly for domain-specific queries, can lead to the amplification of these biases in the retrieved results, which can impact downstream applications.

\section*{Acknowledgement}
This research work has been sponsored by Ant Group Security and Risk Management Fund, the Shanghai Science and Technology Commission Blockchain Special Project (No. 24BC3200100) and the National Key Research and Development Program of China (No. 2023ZD0121402).

\bibliography{custom}

\begin{thebibliography}{51}
\providecommand{\natexlab}[1]{#1}

\bibitem[{Asai et~al.(2022)Asai, Gardner, and
  Hajishirzi}]{DBLP:conf/naacl/Asai0H22}
Akari Asai, Matt Gardner, and Hannaneh Hajishirzi. 2022.
\newblock \href {https://doi.org/10.18653/V1/2022.NAACL-MAIN.162}
  {Evidentiality-guided generation for knowledge-intensive {NLP} tasks}.
\newblock In \emph{Proceedings of the 2022 Conference of the North American
  Chapter of the Association for Computational Linguistics: Human Language
  Technologies, {NAACL} 2022, Seattle, WA, United States, July 10-15, 2022},
  pages 2226--2243. Association for Computational Linguistics.

\bibitem[{Borgeaud et~al.(2022)Borgeaud, Mensch, Hoffmann, Cai, Rutherford,
  Millican, van~den Driessche, Lespiau, Damoc, Clark, de~Las~Casas, Guy,
  Menick, Ring, Hennigan, Huang, Maggiore, Jones, Cassirer, Brock, Paganini,
  Irving, Vinyals, Osindero, Simonyan, Rae, Elsen, and
  Sifre}]{DBLP:conf/icml/BorgeaudMHCRM0L22}
Sebastian Borgeaud, Arthur Mensch, Jordan Hoffmann, Trevor Cai, Eliza
  Rutherford, Katie Millican, George van~den Driessche, Jean{-}Baptiste
  Lespiau, Bogdan Damoc, Aidan Clark, Diego de~Las~Casas, Aurelia Guy, Jacob
  Menick, Roman Ring, Tom Hennigan, Saffron Huang, Loren Maggiore, Chris Jones,
  Albin Cassirer, Andy Brock, Michela Paganini, Geoffrey Irving, Oriol Vinyals,
  Simon Osindero, Karen Simonyan, Jack~W. Rae, Erich Elsen, and Laurent Sifre.
  2022.
\newblock \href {https://proceedings.mlr.press/v162/borgeaud22a.html}
  {Improving language models by retrieving from trillions of tokens}.
\newblock In \emph{International Conference on Machine Learning, {ICML} 2022,
  17-23 July 2022, Baltimore, Maryland, {USA}}, volume 162 of \emph{Proceedings
  of Machine Learning Research}, pages 2206--2240. {PMLR}.

\bibitem[{Chen et~al.(2018)Chen, Song, Wainwright, and
  Jordan}]{DBLP:conf/icml/ChenSWJ18}
Jianbo Chen, Le~Song, Martin~J. Wainwright, and Michael~I. Jordan. 2018.
\newblock \href {http://proceedings.mlr.press/v80/chen18j.html} {Learning to
  explain: An information-theoretic perspective on model interpretation}.
\newblock In \emph{Proceedings of the 35th International Conference on Machine
  Learning, {ICML} 2018, Stockholmsm{\"{a}}ssan, Stockholm, Sweden, July 10-15,
  2018}, volume~80 of \emph{Proceedings of Machine Learning Research}, pages
  882--891. {PMLR}.

\bibitem[{Choi et~al.(2024)Choi, Lee, and Lee}]{DBLP:conf/naacl/ChoiLL24}
Eunseong Choi, Hyeri Lee, and Jongwuk Lee. 2024.
\newblock \href {https://doi.org/10.18653/V1/2024.FINDINGS-NAACL.142}
  {Multi-granularity guided fusion-in-decoder}.
\newblock In \emph{Findings of the Association for Computational Linguistics:
  {NAACL} 2024, Mexico City, Mexico, June 16-21, 2024}, pages 2201--2212.
  Association for Computational Linguistics.

\bibitem[{de~Jong et~al.(2023)de~Jong, Zemlyanskiy, Ainslie, FitzGerald,
  Sanghai, Sha, and Cohen}]{DBLP:conf/acl/JongZAFSSC23}
Michiel de~Jong, Yury Zemlyanskiy, Joshua Ainslie, Nicholas FitzGerald, Sumit
  Sanghai, Fei Sha, and William~W. Cohen. 2023.
\newblock \href {https://doi.org/10.18653/V1/2023.FINDINGS-ACL.732} {Fido:
  Fusion-in-decoder optimized for stronger performance and faster inference}.
\newblock In \emph{Findings of the Association for Computational Linguistics:
  {ACL} 2023, Toronto, Canada, July 9-14, 2023}, pages 11534--11547.
  Association for Computational Linguistics.

\bibitem[{Dong et~al.(2024)Dong, Fatemi, Perozzi, Yang, and
  Tsitsulin}]{DBLP:journals/corr/abs-2405-18414}
Jialin Dong, Bahare Fatemi, Bryan Perozzi, Lin~F. Yang, and Anton Tsitsulin.
  2024.
\newblock \href {https://doi.org/10.48550/ARXIV.2405.18414} {Don't forget to
  connect! improving {RAG} with graph-based reranking}.
\newblock \emph{CoRR}, abs/2405.18414.

\bibitem[{Fan et~al.(2021)Fan, Gong, Liu, Wei, Wang, Jiao, Duan, Zhang, and
  Huang}]{DBLP:conf/naacl/FanGLWWJDZH21}
Zhihao Fan, Yeyun Gong, Dayiheng Liu, Zhongyu Wei, Siyuan Wang, Jian Jiao, Nan
  Duan, Ruofei Zhang, and Xuanjing Huang. 2021.
\newblock \href {https://doi.org/10.18653/V1/2021.NAACL-MAIN.135} {Mask
  attention networks: Rethinking and strengthen transformer}.
\newblock In \emph{Proceedings of the 2021 Conference of the North American
  Chapter of the Association for Computational Linguistics: Human Language
  Technologies, {NAACL-HLT} 2021, Online, June 6-11, 2021}, pages 1692--1701.
  Association for Computational Linguistics.

\bibitem[{Fang et~al.(2024)Fang, Yin, Muralidharan, Heinrich, Pool, Kautz,
  Molchanov, and Wang}]{DBLP:conf/nips/FangYMHPK0W24}
Gongfan Fang, Hongxu Yin, Saurav Muralidharan, Greg Heinrich, Jeff Pool, Jan
  Kautz, Pavlo Molchanov, and Xinchao Wang. 2024.
\newblock \href
  {http://papers.nips.cc/paper\_files/paper/2024/hash/0e9a05f5ce62284c91e4a33498899124-Abstract-Conference.html}
  {Maskllm: Learnable semi-structured sparsity for large language models}.
\newblock In \emph{Advances in Neural Information Processing Systems 38: Annual
  Conference on Neural Information Processing Systems 2024, NeurIPS 2024,
  Vancouver, BC, Canada, December 10 - 15, 2024}.

\bibitem[{Glass et~al.(2022)Glass, Rossiello, Chowdhury, Naik, Cai, and
  Gliozzo}]{DBLP:conf/naacl/GlassRCNCG22}
Michael~R. Glass, Gaetano Rossiello, Md. Faisal~Mahbub Chowdhury, Ankita Naik,
  Pengshan Cai, and Alfio Gliozzo. 2022.
\newblock \href {https://doi.org/10.18653/V1/2022.NAACL-MAIN.194} {Re2g:
  Retrieve, rerank, generate}.
\newblock In \emph{Proceedings of the 2022 Conference of the North American
  Chapter of the Association for Computational Linguistics: Human Language
  Technologies, {NAACL} 2022, Seattle, WA, United States, July 10-15, 2022},
  pages 2701--2715. Association for Computational Linguistics.

\bibitem[{Guu et~al.(2020)Guu, Lee, Tung, Pasupat, and
  Chang}]{DBLP:conf/icml/GuuLTPC20}
Kelvin Guu, Kenton Lee, Zora Tung, Panupong Pasupat, and Ming{-}Wei Chang.
  2020.
\newblock \href {http://proceedings.mlr.press/v119/guu20a.html} {Retrieval
  augmented language model pre-training}.
\newblock In \emph{Proceedings of the 37th International Conference on Machine
  Learning, {ICML} 2020, 13-18 July 2020, Virtual Event}, volume 119 of
  \emph{Proceedings of Machine Learning Research}, pages 3929--3938. {PMLR}.

\bibitem[{Ho et~al.(2020)Ho, Nguyen, Sugawara, and
  Aizawa}]{DBLP:conf/coling/HoNSA20}
Xanh Ho, Anh{-}Khoa~Duong Nguyen, Saku Sugawara, and Akiko Aizawa. 2020.
\newblock \href {https://doi.org/10.18653/V1/2020.COLING-MAIN.580}
  {Constructing {A} multi-hop {QA} dataset for comprehensive evaluation of
  reasoning steps}.
\newblock In \emph{Proceedings of the 28th International Conference on
  Computational Linguistics, {COLING} 2020, Barcelona, Spain (Online), December
  8-13, 2020}, pages 6609--6625. International Committee on Computational
  Linguistics.

\bibitem[{Hofst{\"{a}}tter et~al.(2023)Hofst{\"{a}}tter, Chen, Raman, and
  Zamani}]{DBLP:conf/sigir/HofstatterC0Z23}
Sebastian Hofst{\"{a}}tter, Jiecao Chen, Karthik Raman, and Hamed Zamani. 2023.
\newblock \href {https://doi.org/10.1145/3539618.3591687} {Fid-light: Efficient
  and effective retrieval-augmented text generation}.
\newblock In \emph{Proceedings of the 46th International {ACM} {SIGIR}
  Conference on Research and Development in Information Retrieval, {SIGIR}
  2023, Taipei, Taiwan, July 23-27, 2023}, pages 1437--1447. {ACM}.

\bibitem[{Izacard and Grave(2021{\natexlab{a}})}]{DBLP:conf/iclr/IzacardG21}
Gautier Izacard and Edouard Grave. 2021{\natexlab{a}}.
\newblock \href {https://openreview.net/forum?id=NTEz-6wysdb} {Distilling
  knowledge from reader to retriever for question answering}.
\newblock In \emph{9th International Conference on Learning Representations,
  {ICLR} 2021, Virtual Event, Austria, May 3-7, 2021}. OpenReview.net.

\bibitem[{Izacard and Grave(2021{\natexlab{b}})}]{DBLP:conf/eacl/IzacardG21}
Gautier Izacard and Edouard Grave. 2021{\natexlab{b}}.
\newblock \href {https://doi.org/10.18653/V1/2021.EACL-MAIN.74} {Leveraging
  passage retrieval with generative models for open domain question answering}.
\newblock In \emph{Proceedings of the 16th Conference of the European Chapter
  of the Association for Computational Linguistics: Main Volume, {EACL} 2021,
  Online, April 19 - 23, 2021}, pages 874--880. Association for Computational
  Linguistics.

\bibitem[{Izacard et~al.(2023)Izacard, Lewis, Lomeli, Hosseini, Petroni,
  Schick, Dwivedi{-}Yu, Joulin, Riedel, and
  Grave}]{DBLP:journals/jmlr/IzacardLLHPSDJRG23}
Gautier Izacard, Patrick S.~H. Lewis, Maria Lomeli, Lucas Hosseini, Fabio
  Petroni, Timo Schick, Jane Dwivedi{-}Yu, Armand Joulin, Sebastian Riedel, and
  Edouard Grave. 2023.
\newblock \href {https://jmlr.org/papers/v24/23-0037.html} {Atlas: Few-shot
  learning with retrieval augmented language models}.
\newblock \emph{J. Mach. Learn. Res.}, 24:251:1--251:43.

\bibitem[{Jang et~al.(2017)Jang, Gu, and Poole}]{DBLP:conf/iclr/JangGP17}
Eric Jang, Shixiang Gu, and Ben Poole. 2017.
\newblock \href {https://openreview.net/forum?id=rkE3y85ee} {Categorical
  reparameterization with gumbel-softmax}.
\newblock In \emph{5th International Conference on Learning Representations,
  {ICLR} 2017, Toulon, France, April 24-26, 2017, Conference Track
  Proceedings}. OpenReview.net.

\bibitem[{Karpukhin et~al.(2020)Karpukhin, Oguz, Min, Lewis, Wu, Edunov, Chen,
  and Yih}]{DBLP:conf/emnlp/KarpukhinOMLWEC20}
Vladimir Karpukhin, Barlas Oguz, Sewon Min, Patrick S.~H. Lewis, Ledell Wu,
  Sergey Edunov, Danqi Chen, and Wen{-}tau Yih. 2020.
\newblock \href {https://doi.org/10.18653/V1/2020.EMNLP-MAIN.550} {Dense
  passage retrieval for open-domain question answering}.
\newblock In \emph{Proceedings of the 2020 Conference on Empirical Methods in
  Natural Language Processing, {EMNLP} 2020, Online, November 16-20, 2020},
  pages 6769--6781. Association for Computational Linguistics.

\bibitem[{Ke et~al.(2024)Ke, Kong, Li, Zhang, Mei, and
  Bendersky}]{DBLP:conf/acl/KeK00MB24}
Zixuan Ke, Weize Kong, Cheng Li, Mingyang Zhang, Qiaozhu Mei, and Michael
  Bendersky. 2024.
\newblock \href {https://doi.org/10.18653/V1/2024.ACL-LONG.562} {Bridging the
  preference gap between retrievers and llms}.
\newblock In \emph{Proceedings of the 62nd Annual Meeting of the Association
  for Computational Linguistics (Volume 1: Long Papers), {ACL} 2024, Bangkok,
  Thailand, August 11-16, 2024}, pages 10438--10451. Association for
  Computational Linguistics.

\bibitem[{Khattab and Zaharia(2020)}]{DBLP:conf/sigir/KhattabZ20}
Omar Khattab and Matei Zaharia. 2020.
\newblock \href {https://doi.org/10.1145/3397271.3401075} {Colbert: Efficient
  and effective passage search via contextualized late interaction over
  {BERT}}.
\newblock In \emph{Proceedings of the 43rd International {ACM} {SIGIR}
  conference on research and development in Information Retrieval, {SIGIR}
  2020, Virtual Event, China, July 25-30, 2020}, pages 39--48. {ACM}.

\bibitem[{Kim et~al.(2019)Kim, Kim, and Kwak}]{DBLP:conf/acl/KimKK19}
Daesik Kim, Seonhoon Kim, and Nojun Kwak. 2019.
\newblock \href {https://doi.org/10.18653/V1/P19-1347} {Textbook question
  answering with multi-modal context graph understanding and self-supervised
  open-set comprehension}.
\newblock In \emph{Proceedings of the 57th Conference of the Association for
  Computational Linguistics, {ACL} 2019, Florence, Italy, July 28- August 2,
  2019, Volume 1: Long Papers}, pages 3568--3584. Association for Computational
  Linguistics.

\bibitem[{Kwiatkowski et~al.(2019)Kwiatkowski, Palomaki, Redfield, Collins,
  Parikh, Alberti, Epstein, Polosukhin, Devlin, Lee, Toutanova, Jones, Kelcey,
  Chang, Dai, Uszkoreit, Le, and Petrov}]{DBLP:journals/tacl/KwiatkowskiPRCP19}
Tom Kwiatkowski, Jennimaria Palomaki, Olivia Redfield, Michael Collins,
  Ankur~P. Parikh, Chris Alberti, Danielle Epstein, Illia Polosukhin, Jacob
  Devlin, Kenton Lee, Kristina Toutanova, Llion Jones, Matthew Kelcey,
  Ming{-}Wei Chang, Andrew~M. Dai, Jakob Uszkoreit, Quoc Le, and Slav Petrov.
  2019.
\newblock \href {https://doi.org/10.1162/TACL\_A\_00276} {Natural questions: a
  benchmark for question answering research}.
\newblock \emph{Trans. Assoc. Comput. Linguistics}, 7:452--466.

\bibitem[{Lee et~al.(2019)Lee, Chang, and Toutanova}]{DBLP:conf/acl/LeeCT19}
Kenton Lee, Ming{-}Wei Chang, and Kristina Toutanova. 2019.
\newblock \href {https://doi.org/10.18653/V1/P19-1612} {Latent retrieval for
  weakly supervised open domain question answering}.
\newblock In \emph{Proceedings of the 57th Conference of the Association for
  Computational Linguistics, {ACL} 2019, Florence, Italy, July 28- August 2,
  2019, Volume 1: Long Papers}, pages 6086--6096. Association for Computational
  Linguistics.

\bibitem[{Lewis et~al.(2020)Lewis, Perez, Piktus, Petroni, Karpukhin, Goyal,
  K{\"{u}}ttler, Lewis, Yih, Rockt{\"{a}}schel, Riedel, and
  Kiela}]{DBLP:conf/nips/LewisPPPKGKLYR020}
Patrick S.~H. Lewis, Ethan Perez, Aleksandra Piktus, Fabio Petroni, Vladimir
  Karpukhin, Naman Goyal, Heinrich K{\"{u}}ttler, Mike Lewis, Wen{-}tau Yih,
  Tim Rockt{\"{a}}schel, Sebastian Riedel, and Douwe Kiela. 2020.
\newblock \href
  {https://proceedings.neurips.cc/paper/2020/hash/6b493230205f780e1bc26945df7481e5-Abstract.html}
  {Retrieval-augmented generation for knowledge-intensive {NLP} tasks}.
\newblock In \emph{Advances in Neural Information Processing Systems 33: Annual
  Conference on Neural Information Processing Systems 2020, NeurIPS 2020,
  December 6-12, 2020, virtual}.

\bibitem[{Li et~al.(2023)Li, Lei, and Yang}]{DBLP:conf/icassp/LiLY23}
Xin{-}Yi Li, Wei{-}Jun Lei, and Yu{-}Bin Yang. 2023.
\newblock \href {https://doi.org/10.1109/ICASSP49357.2023.10096119} {From easy
  to hard: Two-stage selector and reader for multi-hop question answering}.
\newblock In \emph{{IEEE} International Conference on Acoustics, Speech and
  Signal Processing {ICASSP} 2023, Rhodes Island, Greece, June 4-10, 2023},
  pages 1--5. {IEEE}.

\bibitem[{Li et~al.(2024)Li, Zhang, Pan, Sun, Duan, Fang, Han, Wang, and
  Wang}]{DBLP:journals/corr/abs-2408-11745}
Zhenyu Li, Yike Zhang, Tengyu Pan, Yutao Sun, Zhichao Duan, Junjie Fang, Rong
  Han, Zixuan Wang, and Jianyong Wang. 2024.
\newblock \href {https://doi.org/10.48550/ARXIV.2408.11745} {Focusllm: Scaling
  llm's context by parallel decoding}.
\newblock \emph{CoRR}, abs/2408.11745.

\bibitem[{Lin et~al.(2024)Lin, Chen, Chen, Shi, Lomeli, James, Rodriguez, Kahn,
  Szilvasy, Lewis, Zettlemoyer, and Yih}]{DBLP:conf/iclr/Lin0CSL00KSLZY24}
Xi~Victoria Lin, Xilun Chen, Mingda Chen, Weijia Shi, Maria Lomeli, Richard
  James, Pedro Rodriguez, Jacob Kahn, Gergely Szilvasy, Mike Lewis, Luke
  Zettlemoyer, and Wen{-}tau Yih. 2024.
\newblock \href {https://openreview.net/forum?id=22OTbutug9} {{RA-DIT:}
  retrieval-augmented dual instruction tuning}.
\newblock In \emph{The Twelfth International Conference on Learning
  Representations, {ICLR} 2024, Vienna, Austria, May 7-11, 2024}.
  OpenReview.net.

\bibitem[{Raffel et~al.(2020)Raffel, Shazeer, Roberts, Lee, Narang, Matena,
  Zhou, Li, and Liu}]{DBLP:journals/jmlr/RaffelSRLNMZLL20}
Colin Raffel, Noam Shazeer, Adam Roberts, Katherine Lee, Sharan Narang, Michael
  Matena, Yanqi Zhou, Wei Li, and Peter~J. Liu. 2020.
\newblock \href {https://jmlr.org/papers/v21/20-074.html} {Exploring the limits
  of transfer learning with a unified text-to-text transformer}.
\newblock \emph{J. Mach. Learn. Res.}, 21:140:1--140:67.

\bibitem[{Ratner et~al.(2023)Ratner, Levine, Belinkov, Ram, Magar, Abend,
  Karpas, Shashua, Leyton{-}Brown, and
  Shoham}]{DBLP:conf/acl/RatnerLBRMAKSLS23}
Nir Ratner, Yoav Levine, Yonatan Belinkov, Ori Ram, Inbal Magar, Omri Abend,
  Ehud Karpas, Amnon Shashua, Kevin Leyton{-}Brown, and Yoav Shoham. 2023.
\newblock \href {https://doi.org/10.18653/V1/2023.ACL-LONG.352} {Parallel
  context windows for large language models}.
\newblock In \emph{Proceedings of the 61st Annual Meeting of the Association
  for Computational Linguistics (Volume 1: Long Papers), {ACL} 2023, Toronto,
  Canada, July 9-14, 2023}, pages 6383--6402. Association for Computational
  Linguistics.

\bibitem[{Robertson and Zaragoza(2009)}]{DBLP:journals/ftir/RobertsonZ09}
Stephen~E. Robertson and Hugo Zaragoza. 2009.
\newblock \href {https://doi.org/10.1561/1500000019} {The probabilistic
  relevance framework: {BM25} and beyond}.
\newblock \emph{Found. Trends Inf. Retr.}, 3(4):333--389.

\bibitem[{Sachan et~al.(2023)Sachan, Lewis, Yogatama, Zettlemoyer, Pineau, and
  Zaheer}]{DBLP:journals/tacl/SachanLYZPZ23}
Devendra~Singh Sachan, Mike Lewis, Dani Yogatama, Luke Zettlemoyer, Joelle
  Pineau, and Manzil Zaheer. 2023.
\newblock \href {https://doi.org/10.1162/TACL\_A\_00564} {Questions are all you
  need to train a dense passage retriever}.
\newblock \emph{Trans. Assoc. Comput. Linguistics}, 11:600--616.

\bibitem[{Sachan et~al.(2021)Sachan, Reddy, Hamilton, Dyer, and
  Yogatama}]{DBLP:conf/nips/SachanRHDY21}
Devendra~Singh Sachan, Siva Reddy, William~L. Hamilton, Chris Dyer, and Dani
  Yogatama. 2021.
\newblock \href
  {https://proceedings.neurips.cc/paper/2021/hash/da3fde159d754a2555eaa198d2d105b2-Abstract.html}
  {End-to-end training of multi-document reader and retriever for open-domain
  question answering}.
\newblock In \emph{Advances in Neural Information Processing Systems 34: Annual
  Conference on Neural Information Processing Systems 2021, NeurIPS 2021,
  December 6-14, 2021, virtual}, pages 25968--25981.

\bibitem[{Salemi and Zamani(2024)}]{DBLP:conf/sigir/SalemiZ24a}
Alireza Salemi and Hamed Zamani. 2024.
\newblock \href {https://doi.org/10.1145/3626772.3657957} {Evaluating retrieval
  quality in retrieval-augmented generation}.
\newblock In \emph{Proceedings of the 47th International {ACM} {SIGIR}
  Conference on Research and Development in Information Retrieval, {SIGIR}
  2024, Washington DC, USA, July 14-18, 2024}, pages 2395--2400. {ACM}.

\bibitem[{Shi et~al.(2024)Shi, Min, Yasunaga, Seo, James, Lewis, Zettlemoyer,
  and Yih}]{DBLP:conf/naacl/ShiMYS0LZY24}
Weijia Shi, Sewon Min, Michihiro Yasunaga, Minjoon Seo, Richard James, Mike
  Lewis, Luke Zettlemoyer, and Wen{-}tau Yih. 2024.
\newblock \href {https://doi.org/10.18653/V1/2024.NAACL-LONG.463} {{REPLUG:}
  retrieval-augmented black-box language models}.
\newblock In \emph{Proceedings of the 2024 Conference of the North American
  Chapter of the Association for Computational Linguistics: Human Language
  Technologies (Volume 1: Long Papers), {NAACL} 2024, Mexico City, Mexico, June
  16-21, 2024}, pages 8371--8384. Association for Computational Linguistics.

\bibitem[{Tang and Yang(2024)}]{DBLP:journals/corr/abs-2401-15391}
Yixuan Tang and Yi~Yang. 2024.
\newblock \href {https://doi.org/10.48550/ARXIV.2401.15391} {Multihop-rag:
  Benchmarking retrieval-augmented generation for multi-hop queries}.
\newblock \emph{CoRR}, abs/2401.15391.

\bibitem[{Trivedi et~al.(2022)Trivedi, Balasubramanian, Khot, and
  Sabharwal}]{DBLP:journals/tacl/TrivediBKS22}
Harsh Trivedi, Niranjan Balasubramanian, Tushar Khot, and Ashish Sabharwal.
  2022.
\newblock \href {https://doi.org/10.1162/TACL\_A\_00475} {Musique: Multihop
  questions via single-hop question composition}.
\newblock \emph{Trans. Assoc. Comput. Linguistics}, 10:539--554.

\bibitem[{Tu et~al.(2020)Tu, Huang, Wang, Huang, He, and
  Zhou}]{DBLP:conf/aaai/TuHW0HZ20}
Ming Tu, Kevin Huang, Guangtao Wang, Jing Huang, Xiaodong He, and Bowen Zhou.
  2020.
\newblock \href {https://doi.org/10.1609/AAAI.V34I05.6441} {Select, answer and
  explain: Interpretable multi-hop reading comprehension over multiple
  documents}.
\newblock In \emph{The Thirty-Fourth {AAAI} Conference on Artificial
  Intelligence, {AAAI} 2020, The Thirty-Second Innovative Applications of
  Artificial Intelligence Conference, {IAAI} 2020, The Tenth {AAAI} Symposium
  on Educational Advances in Artificial Intelligence, {EAAI} 2020, New York,
  NY, USA, February 7-12, 2020}, pages 9073--9080. {AAAI} Press.

\bibitem[{Wang et~al.(2023)Wang, Yu, and Zhang}]{DBLP:conf/acl/WangY023}
Cunxiang Wang, Haofei Yu, and Yue Zhang. 2023.
\newblock \href {https://doi.org/10.18653/V1/2023.FINDINGS-ACL.155} {Rfid:
  Towards rational fusion-in-decoder for open-domain question answering}.
\newblock In \emph{Findings of the Association for Computational Linguistics:
  {ACL} 2023, Toronto, Canada, July 9-14, 2023}, pages 2473--2481. Association
  for Computational Linguistics.

\bibitem[{Wu et~al.(2021)Wu, Zhang, and
  Zhao}]{DBLP:journals/corr/abs-2107-11823}
Bohong Wu, Zhuosheng Zhang, and Hai Zhao. 2021.
\newblock \href {https://arxiv.org/abs/2107.11823} {Graph-free multi-hop
  reading comprehension: {A} select-to-guide strategy}.
\newblock \emph{CoRR}, abs/2107.11823.

\bibitem[{Xiao and Liu(2023)}]{xiao2023bge}
Shitao Xiao and Zheng Liu. 2023.
\newblock \href {https://github.com/FlagOpen/FlagEmbedding} {Baai general
  embedding}.

\bibitem[{Xie and Ermon(2019)}]{DBLP:conf/ijcai/XieE19}
Sang~Michael Xie and Stefano Ermon. 2019.
\newblock \href {https://doi.org/10.24963/IJCAI.2019/544} {Reparameterizable
  subset sampling via continuous relaxations}.
\newblock In \emph{Proceedings of the Twenty-Eighth International Joint
  Conference on Artificial Intelligence, {IJCAI} 2019, Macao, China, August
  10-16, 2019}, pages 3919--3925. ijcai.org.

\bibitem[{Xie et~al.(2020)Xie, Dai, Chen, Dai, Zhao, Zha, Wei, and
  Pfister}]{DBLP:conf/nips/XieDCDZZ0P20}
Yujia Xie, Hanjun Dai, Minshuo Chen, Bo~Dai, Tuo Zhao, Hongyuan Zha, Wei Wei,
  and Tomas Pfister. 2020.
\newblock \href
  {https://proceedings.neurips.cc/paper/2020/hash/ec24a54d62ce57ba93a531b460fa8d18-Abstract.html}
  {Differentiable top-k with optimal transport}.
\newblock In \emph{Advances in Neural Information Processing Systems 33: Annual
  Conference on Neural Information Processing Systems 2020, NeurIPS 2020,
  December 6-12, 2020, virtual}.

\bibitem[{Xiong et~al.(2021)Xiong, Li, Iyer, Du, Lewis, Wang, Mehdad, Yih,
  Riedel, Kiela, and Oguz}]{DBLP:conf/iclr/XiongLIDLWMY0KO21}
Wenhan Xiong, Xiang~Lorraine Li, Srini Iyer, Jingfei Du, Patrick S.~H. Lewis,
  William~Yang Wang, Yashar Mehdad, Scott Yih, Sebastian Riedel, Douwe Kiela,
  and Barlas Oguz. 2021.
\newblock \href {https://openreview.net/forum?id=EMHoBG0avc1} {Answering
  complex open-domain questions with multi-hop dense retrieval}.
\newblock In \emph{9th International Conference on Learning Representations,
  {ICLR} 2021, Virtual Event, Austria, May 3-7, 2021}. OpenReview.net.

\bibitem[{Yang et~al.(2019)Yang, Zhang, Ni, Li, Liu, Zhou, and
  Tian}]{DBLP:conf/cvpr/YangZNLLZT19}
Jiancheng Yang, Qiang Zhang, Bingbing Ni, Linguo Li, Jinxian Liu, Mengdie Zhou,
  and Qi~Tian. 2019.
\newblock \href {https://doi.org/10.1109/CVPR.2019.00344} {Modeling point
  clouds with self-attention and gumbel subset sampling}.
\newblock In \emph{{IEEE} Conference on Computer Vision and Pattern
  Recognition, {CVPR} 2019, Long Beach, CA, USA, June 16-20, 2019}, pages
  3323--3332. Computer Vision Foundation / {IEEE}.

\bibitem[{Yang et~al.(2018)Yang, Qi, Zhang, Bengio, Cohen, Salakhutdinov, and
  Manning}]{DBLP:conf/emnlp/Yang0ZBCSM18}
Zhilin Yang, Peng Qi, Saizheng Zhang, Yoshua Bengio, William~W. Cohen, Ruslan
  Salakhutdinov, and Christopher~D. Manning. 2018.
\newblock \href {https://doi.org/10.18653/V1/D18-1259} {Hotpotqa: {A} dataset
  for diverse, explainable multi-hop question answering}.
\newblock In \emph{Proceedings of the 2018 Conference on Empirical Methods in
  Natural Language Processing, Brussels, Belgium, October 31 - November 4,
  2018}, pages 2369--2380. Association for Computational Linguistics.

\bibitem[{Yen et~al.(2024)Yen, Gao, and Chen}]{DBLP:conf/acl/YenG024}
Howard Yen, Tianyu Gao, and Danqi Chen. 2024.
\newblock \href {https://doi.org/10.18653/V1/2024.ACL-LONG.142} {Long-context
  language modeling with parallel context encoding}.
\newblock In \emph{Proceedings of the 62nd Annual Meeting of the Association
  for Computational Linguistics (Volume 1: Long Papers), {ACL} 2024, Bangkok,
  Thailand, August 11-16, 2024}, pages 2588--2610. Association for
  Computational Linguistics.

\bibitem[{Yin et~al.(2023)Yin, Wang, Hu, Wu, Yan, Zhang, Cao, Huang, and
  Qiu}]{DBLP:conf/cncl/YinWHWYZCHQ23}
Zhangyue Yin, Yuxin Wang, Xiannian Hu, Yiguang Wu, Hang Yan, Xinyu Zhang, Zhao
  Cao, Xuanjing Huang, and Xipeng Qiu. 2023.
\newblock \href {https://doi.org/10.1007/978-981-99-6207-5\_5} {Rethinking
  label smoothing on multi-hop question answering}.
\newblock In \emph{Chinese Computational Linguistics - 22nd China National
  Conference, {CCL} 2023, Harbin, China, August 3-5, 2023, Proceedings}, volume
  14232 of \emph{Lecture Notes in Computer Science}, pages 72--87. Springer.

\bibitem[{Yu et~al.(2022)Yu, Zhu, Fang, Yu, Wang, Xu, Ren, Yang, and
  Zeng}]{DBLP:conf/acl/Yu0F0WXRY022}
Donghan Yu, Chenguang Zhu, Yuwei Fang, Wenhao Yu, Shuohang Wang, Yichong Xu,
  Xiang Ren, Yiming Yang, and Michael Zeng. 2022.
\newblock \href {https://doi.org/10.18653/V1/2022.ACL-LONG.340} {Kg-fid:
  Infusing knowledge graph in fusion-in-decoder for open-domain question
  answering}.
\newblock In \emph{Proceedings of the 60th Annual Meeting of the Association
  for Computational Linguistics (Volume 1: Long Papers), {ACL} 2022, Dublin,
  Ireland, May 22-27, 2022}, pages 4961--4974. Association for Computational
  Linguistics.

\bibitem[{Zhang et~al.(2024)Zhang, Zhang, Zhang, Liu, and
  Huang}]{DBLP:conf/naacl/ZhangZ00H24}
Jiahao Zhang, Haiyang Zhang, Dongmei Zhang, Yong Liu, and Shen Huang. 2024.
\newblock \href {https://doi.org/10.18653/V1/2024.NAACL-LONG.96} {End-to-end
  beam retrieval for multi-hop question answering}.
\newblock In \emph{Proceedings of the 2024 Conference of the North American
  Chapter of the Association for Computational Linguistics: Human Language
  Technologies (Volume 1: Long Papers), {NAACL} 2024, Mexico City, Mexico, June
  16-21, 2024}, pages 1718--1731. Association for Computational Linguistics.

\bibitem[{Zhao et~al.(2021)Zhao, Xiong, Boyd{-}Graber, and
  III}]{DBLP:conf/naacl/ZhaoXBD21}
Chen Zhao, Chenyan Xiong, Jordan~L. Boyd{-}Graber, and Hal~Daum{\'{e}} III.
  2021.
\newblock \href {https://doi.org/10.18653/V1/2021.NAACL-MAIN.368} {Multi-step
  reasoning over unstructured text with beam dense retrieval}.
\newblock In \emph{Proceedings of the 2021 Conference of the North American
  Chapter of the Association for Computational Linguistics: Human Language
  Technologies, {NAACL-HLT} 2021, Online, June 6-11, 2021}, pages 4635--4641.
  Association for Computational Linguistics.

\bibitem[{Zhu et~al.(2021)Zhu, Lei, Wang, Zheng, Poria, and
  Chua}]{DBLP:journals/corr/abs-2101-00774}
Fengbin Zhu, Wenqiang Lei, Chao Wang, Jianming Zheng, Soujanya Poria, and
  Tat{-}Seng Chua. 2021.
\newblock \href {https://arxiv.org/abs/2101.00774} {Retrieving and reading: {A}
  comprehensive survey on open-domain question answering}.
\newblock \emph{CoRR}, abs/2101.00774.

\bibitem[{Zhuang et~al.(2023)Zhuang, Qin, Jagerman, Hui, Ma, Lu, Ni, Wang, and
  Bendersky}]{DBLP:conf/sigir/Zhuang0J0MLNWB23}
Honglei Zhuang, Zhen Qin, Rolf Jagerman, Kai Hui, Ji~Ma, Jing Lu, Jianmo Ni,
  Xuanhui Wang, and Michael Bendersky. 2023.
\newblock \href {https://doi.org/10.1145/3539618.3592047} {Rankt5: Fine-tuning
  {T5} for text ranking with ranking losses}.
\newblock In \emph{Proceedings of the 46th International {ACM} {SIGIR}
  Conference on Research and Development in Information Retrieval, {SIGIR}
  2023, Taipei, Taiwan, July 23-27, 2023}, pages 2308--2313. {ACM}.

\end{thebibliography}

\appendix
\newpage
\begin{figure*}[!th]
  \includegraphics[width=\linewidth]{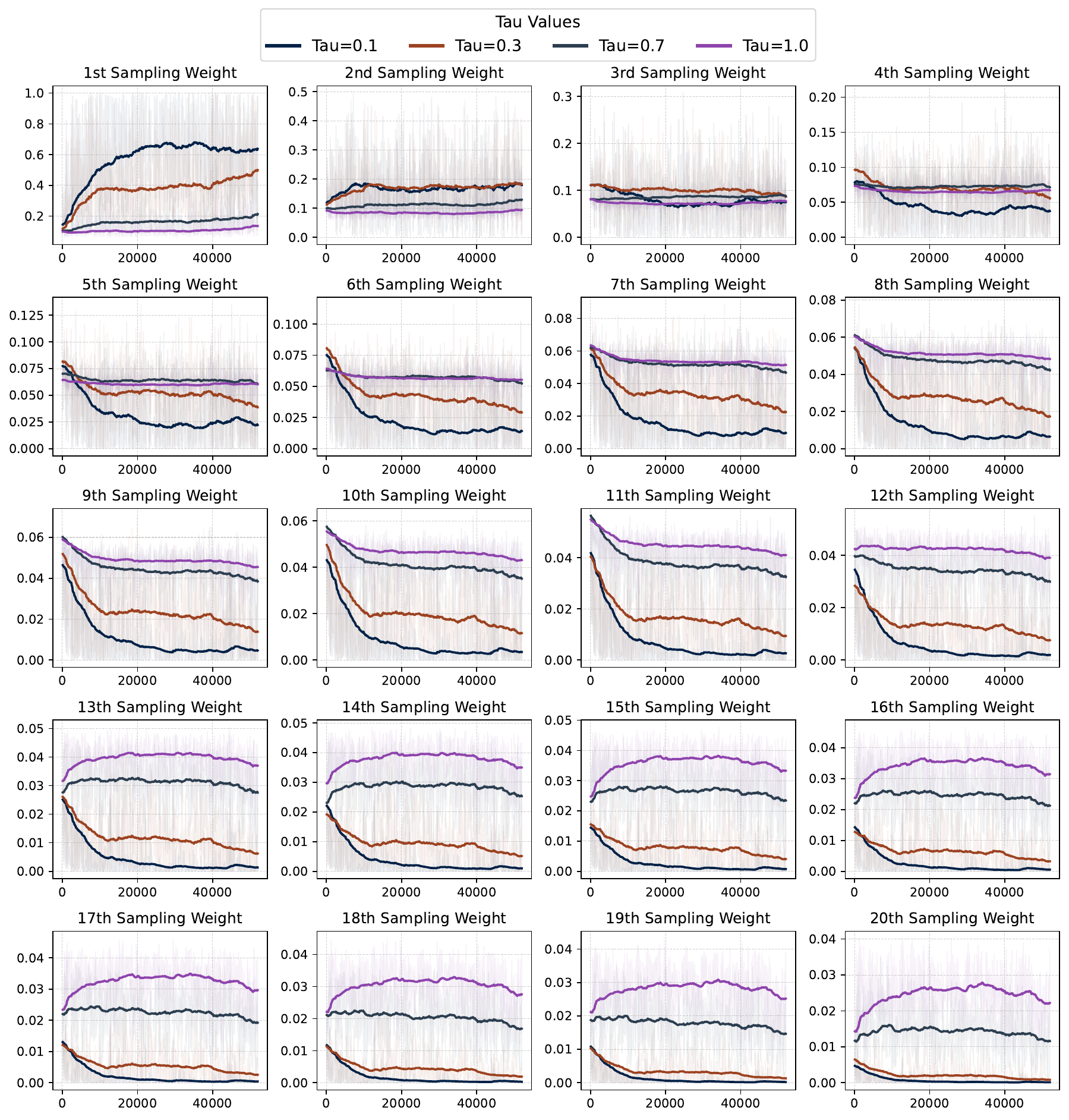}
  \centering
\caption{The impact of different \(\tau\) values on the training process. We conduct with 20 candidate documents and RankT5 on the NQ dataset. The solid line in the figure represents the moving average. The differences in sampling weights indicate the Reranker's ability to distinguish between candidate documents.}
\label{fig:all_tau_sampling_weight}
\end{figure*}

\begin{figure*}[!th]
  \includegraphics[width=\linewidth]{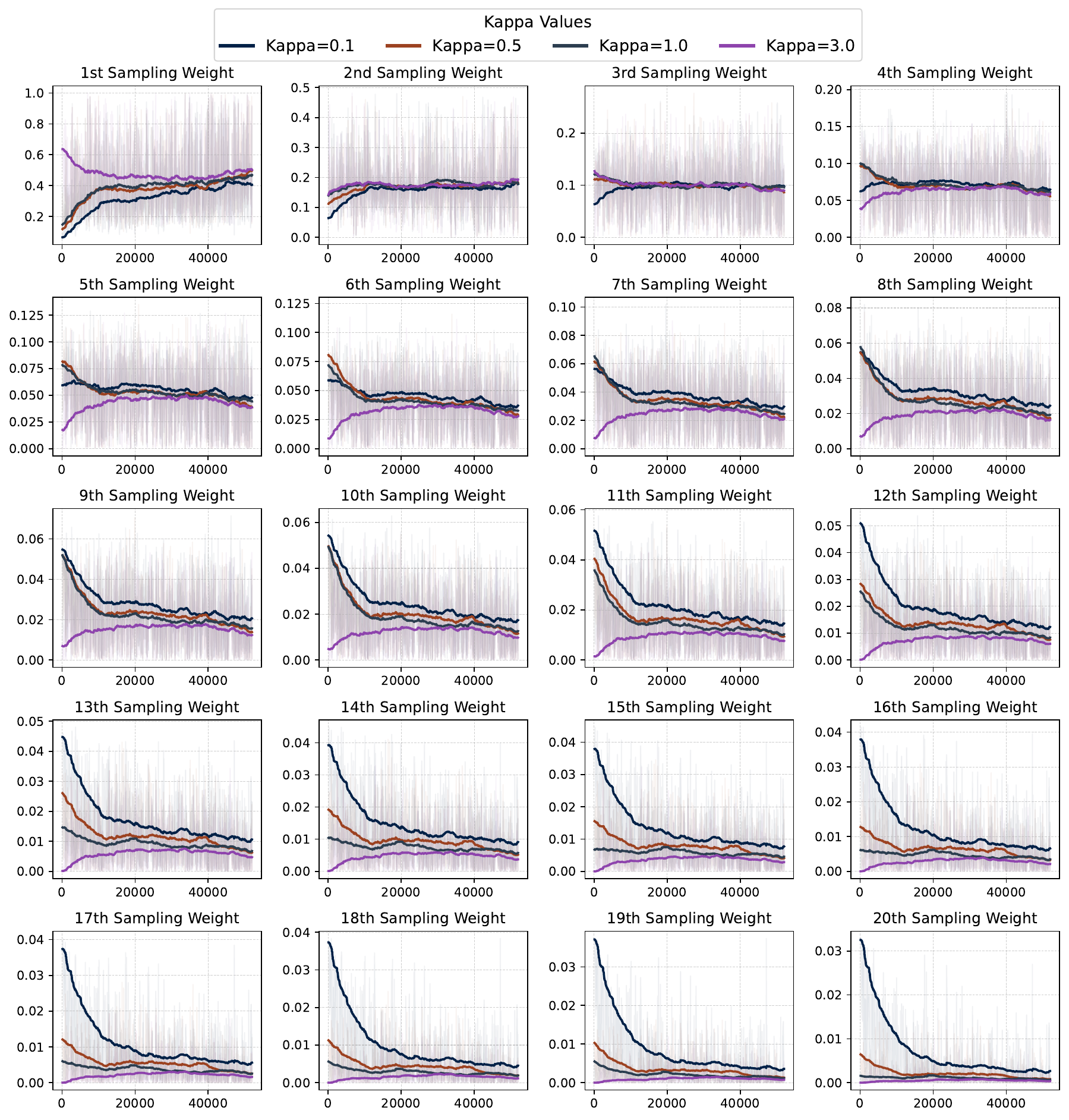}
  \centering
\caption{The impact of different \(\kappa\) values on the training process. We conduct with 20 candidate documents and RankT5 on the NQ dataset. The solid line in the figure represents the moving average. The differences in sampling weights indicate the Reranker's ability to distinguish between candidate documents.}
\label{fig:all_kappa_sampling_weight}
\end{figure*}

\section{Effect of hyper-parameters on the Training Process}\label{sec: Effect of hyper-parameters on the Training Process}
The temperature parameter \(\tau\) controls the sharpness of the softmax distribution used in the selection process of documents. We conduct sampling weight learning for 20 candidate documents based on RankT5 on the NQ dataset, and tested the impact of different \(\tau\) values on the sampling weights during the training process. The experimental results are shown in \autoref{fig:all_tau_sampling_weight}. Specifically, as \(\tau\) approaches zero, the softmax distribution becomes increasingly sharp, leading to a \textit{hard} selection process where the model heavily favors the document with the highest score. This results in a deterministic decision-making process, where the model’s focus is on exploitation, quickly converging to a particular document. On the other hand, when \(\tau\) increases, the distribution becomes smoother, allowing for a more stochastic sampling process. This introduces more exploration, as the model is less likely to fixate on a single document, encouraging the exploration of other potential candidates. A larger \(\tau\) thus promotes diversity in the selection process, which can be beneficial for avoiding local optima and improving generalization during training.

The scaling factor \(\kappa\) plays a critical role in controlling the relative influence of the Reranker scores on the overall document selection process. We test the impact of different \(\kappa\) values on the sampling weights during the training process. The experimental results are shown in \autoref{fig:all_kappa_sampling_weight}. Specifically, \(\kappa\) modulates the contribution of the Reranker score \(w_i\) to the final selection probability. When \(\kappa\) is small, the contribution of the original Gumbel noise term \(G_i\) dominates the selection process. This introduces significant randomness, increasing the exploration rate during training. A small \(\kappa\) value results in noisy selection, encouraging the model to explore various documents and learn more diverse representations. Conversely, when \(\kappa\) is large, the Reranker score \(w_i\) has a stronger influence, and the model’s selection becomes more deterministic. In this case, the Reranker score dominates the sampling process, leading to faster convergence as the model focuses on selecting the most highly scored documents. However, an overly large \(\kappa\) may limit the exploration of alternative options, potentially leading to overfitting and reduced generalization.

\section{LLM-Supervised Baselines} \label{sec: appendix_baseline}
\paragraph{Baselines.}
We compare our approach against four LLM-supervised reranker training methods that leverage generative language model signals to supervise retriever learning without requiring additional document annotations. In particular, we consider the following methods:

\textbf{Attention Distillation (ADist)}~\cite{DBLP:conf/iclr/IzacardG21}: This method utilizes the cross-attention scores from the language model—augmented by the norms of the corresponding value vectors—to compute a target relevance distribution over retrieved documents. The reranker $\mathcal{R}$ is trained by minimizing the KL-divergence between its own distribution over the top-$k$ documents and the attention-based target distribution. The target distribution for the reranker is defined as:
\[
p_{\textsc{attn}}(\mathbf{p}_k) = \frac{\exp(\alpha_k \| \mathbf{v}_k \|_2)}{\sum_{i=1}^K \exp(\alpha_i \| \mathbf{v}_i \|_2)}
\]
where $\alpha_k$ is the attention score for document $\mathbf{p}_k$ and $\|\mathbf{v}_k\|_2$ is the L2 norm of the corresponding value vector. The loss function minimizes the KL-divergence between the reranker's distribution $p_{\mathcal{R}}$ and the target distribution $p_{\textsc{attn}}$:
\[
\textsc{KL}(p_{\textsc{attn}} \ \| \ p_{\mathcal{R}}) = \sum_{k=1}^K p_{\textsc{attn}} (\mathbf{p}_k) \log \frac{p_{\textsc{attn}} (\mathbf{p}_k)}{p_{\mathcal{R}} (\mathbf{p}_k)}
\]

\textbf{End-to-end Multi-Document Reader and Reranker (EMDR$^2$)}~\cite{DBLP:conf/nips/SachanRHDY21, DBLP:conf/naacl/ShiMYS0LZY24, DBLP:conf/iclr/Lin0CSL00KSLZY24}: EMDR$^2$ adopts an expectation-maximization approach, treating the retrieved documents as latent variables. Given a query $\mathbf{q}$ and a corresponding answer $\mathbf{a}$, along with the top-$k$ retrieved documents, the loss is designed to maximize the log-likelihood of the output given these documents. The objective function is:
\[
\mathcal{L}_{\text{EMDR}^2} = \log \left[ \sum_{k=1}^K p_{\textsc{lm}}(\mathbf{a} \ | \ \mathbf{q}, \mathbf{p}_k) p_{\mathcal{R}}(\mathbf{p}_k \ | \ \mathbf{q}) \right]
\]
where $p_{\textsc{lm}}(\mathbf{a} \ | \ \mathbf{q}, \mathbf{p}_k)$ is the language model’s probability of generating the answer $\mathbf{a}$ conditioned on the query $\mathbf{q}$ and document $\mathbf{p}_k$, and $p_{\mathcal{R}}(\mathbf{p}_k \ | \ \mathbf{q})$ is the reranker’s distribution over the top-$k$ documents.

\textbf{Perplexity Distillation (PDist)}~\cite{DBLP:journals/jmlr/IzacardLLHPSDJRG23, DBLP:conf/naacl/GlassRCNCG22}: In this approach, the reranker is trained to predict the improvement in the language model’s perplexity when each document is used to condition the model’s output. The KL divergence is minimized between the reranker's distribution over documents and the posterior distribution derived from the language model, which provides a direct measure of how much a document contributes to the model’s performance. The target distribution for the reranker is computed as:
\[
p_k = \frac{\exp(\log p_{\textsc{lm}} (\mathbf{a} \ | \ \mathbf{p}_k, \mathbf{q}))}{\sum_{i=1}^K \exp ( \log p_{\textsc{lm}} (\mathbf{a} \ | \ \mathbf{p}_i, \mathbf{q}))}
\]
The reranker is trained to minimize the KL-divergence between its predicted distribution over the documents and this target distribution:

\[
\mathcal{L}_{\text{PDist}} = \sum_{k=1}^K p_{\mathcal{R}}(\mathbf{p}_k \ | \ \mathbf{q}) \log \frac{p_{\mathcal{R}}(\mathbf{p}_k \ | \ \mathbf{q})}{p_k}
\]

Here, $p_k$ is the distribution over documents that the language model would prefer, and the reranker is trained to match this distribution to improve the language model's perplexity. The KL-divergence loss encourages the reranker to select documents that enhance the model's ability to generate the correct answer.

The objective function in EMDR$^2$ is based on maximizing the log-likelihood of the correct answer, given the query and documents. This method treats the documents as latent variables and aims to optimize the likelihood of generating the correct answer based on the combination of the language model and reranker’s distributions. On the other hand, PDist focuses on optimizing the reranker’s distribution by minimizing the KL-divergence between its predictions and the target distribution, which is derived from the language model’s perplexity.

\textbf{Leave-one-out Perplexity Distillation (LOOP)}~\cite{DBLP:journals/jmlr/IzacardLLHPSDJRG23}: LOOP refines the PDist approach by considering the impact of each document in the context of all other documents in the top-$k$ set. For each document, the log-likelihood of the output is computed by excluding the document from the retrieval set, and the negative of this value is used as a relevance score. The target distribution is:
\begin{equation*}
    \begin{aligned}
        &p_{\textsc{loop}}(\mathbf{p}_k) = \\
        &\frac{\exp(- \log p_{\textsc{lm}} (\mathbf{a} \ | \ \mathcal{D}_K \setminus \{ \mathbf{p}_k \}, \mathbf{q}))}
        {\sum_{i=1}^K \exp (- \log p_{\textsc{lm}} (\mathbf{a} \ | \ \mathcal{D}_K \setminus \{ \mathbf{p}_i \}, \mathbf{q}))}
    \end{aligned}
\end{equation*}
The reranker is trained to minimize the KL-divergence between this distribution and the one obtained from the reranker.

\section{More Details}
\subsection{Variants of FiD}\label{sec: appendix_fid}
Recent advancements in Open-Domain Question Answering have led to the development of several enhanced Fusion-in-Decoder models. KG-FiD~\cite{DBLP:conf/acl/Yu0F0WXRY022} enhances the traditional FiD framework by integrating knowledge graphs to establish structural relationships among retrieved passages. This integration employs graph neural networks to re-rank passages, selecting the most pertinent ones for answer generation, thereby improving both effectiveness and efficiency. FiDO~\cite{DBLP:conf/acl/JongZAFSSC23} addresses memory bandwidth constraints inherent in the FiD architecture by reallocating computational resources. This optimization results in a significant increase in inference speed without compromising performance, making it more suitable for real-time applications. FiD-Light~\cite{DBLP:conf/sigir/HofstatterC0Z23} focuses on efficient retrieval-augmented text generation by optimizing the balance between retrieval and generation components. This approach reduces computational overhead while maintaining answer accuracy, offering a more resource-efficient alternative. RFiD~\cite{DBLP:conf/acl/WangY023} introduces a multi-task learning approach to discern evidentiality, combining passage re-ranking with sentence classification. This method enhances the model's ability to identify causal relationships between questions and passages, leading to improved answer accuracy. Multi-Granularity Guided Fusion-in-Decoder (MG-FiD)~\cite{DBLP:conf/naacl/ChoiLL24} further refines the FiD approach by aggregating evidence across multiple levels of granularity. It harmonizes passage re-ranking with sentence-level classification, enhancing both accuracy and decoding efficiency.

In our experiments, since we are mainly focusing on reranker training strategies rather than reader, we utilize the classical Fusion-in-Decoder model architecture. Building upon this foundation, we compare our approach with various LLM-supervised reranker training strategies to assess their impact on ODQA performance.

\begin{table*}[!ht]
\centering
\small
\resizebox{0.9\linewidth}{!}{
\begin{tabular}{ l || c c|| c c | c || c c c }
    \toprule
    & \multicolumn{2}{c||}{\textbf{Mining Setting}} & \multicolumn{3}{c||}{\textbf{Reranker Setting}} & \multicolumn{3}{c}{\textbf{Generator Setting}} \\  
    \cmidrule(r){2-3} \cmidrule(r){4-6} \cmidrule(r){7-9}
    & \textbf{Recall@5} & \textbf{NDCG@5} & \textbf{Recall@5} & \textbf{NDCG@5} & \textbf{MRR} & \textbf{EM} & \textbf{SubEM} & \textbf{F1} \\
    \addlinespace[0.2em]
    \hline
    \hline
    \addlinespace[0.4em]
    \multicolumn{9}{c}{\textbf{Dataset: }\normalsize\texttt{Hotpotqa}} \\[-0.1em]
    \addlinespace[0.1em]\multicolumn{1}{l}{\bf{Reranker: }\normalsize\texttt{Rank-T5}}& \multicolumn{8}{l}{}\\[0.2em]
    - EMDR~\cite{DBLP:conf/iclr/Lin0CSL00KSLZY24}           & \underline{78.8}  & \underline{81.1}  & \underline{79.5} & \underline{81.1} & \underline{95.8} & \underline{58.3} & \underline{64.6} & \underline{73.1} \\
    - PDist~\cite{DBLP:conf/naacl/GlassRCNCG22}         & 70.9  & 73.9  & 71.4 & 73.6 & 92.6 & 57.8 & 64.0 & 72.5 \\
    - LOOP~\cite{DBLP:journals/jmlr/IzacardLLHPSDJRG23}           & 72.2  & 75.4  & 73.4 & 75.7 & 93.7 & 58.0 & 64.2 & 72.7 \\
    - ADist~\cite{DBLP:conf/iclr/IzacardG21}& 72.2  & 74.1  & 72.5 & 73.8 & 90.5 & 56.5 & 62.6 & 71.1 \\
    \rowcolor{gray!10} - G-Rerank   & \bf 81.9  & \bf 83.7  & \bf 83.1 & \bf 84.0 & \bf 95.9 & \bf 58.8 & \bf 65.1 & \bf 73.5 \\
    \midrule
     \addlinespace[0.4em]
     \addlinespace[0.1em]\multicolumn{1}{l}{\bf{Reranker: }\normalsize\texttt{BGE-Base}}& \multicolumn{8}{l}{}\\[0.2em]
    - EMDR~\cite{DBLP:conf/iclr/Lin0CSL00KSLZY24}           & 78.4  & \underline{81.1}  & 78.8 & 80.7 & \underline{95.9} & 58.6 & 64.8 & \underline{73.4} \\
    - PDist~\cite{DBLP:conf/naacl/GlassRCNCG22}         & 76.7  & 79.8  & 78.5 & 80.7 & \bf 96.1 & \underline{58.7} & \underline{64.9} & \underline{73.4} \\
    - LOOP~\cite{DBLP:journals/jmlr/IzacardLLHPSDJRG23}           & 76.0  & 79.0  & 77.1 & 79.2 & 95.3 & 58.3 & 64.5 & 73.0 \\
    - ADist~\cite{DBLP:conf/iclr/IzacardG21}& \underline{78.5}  & 80.7  & \underline{79.3} & \underline{80.9} & 95.1 & 58.2 & 64.4 & 73.0 \\
    \rowcolor{gray!10} - G-Rerank   & \bf 81.6  & \bf 83.3  & \bf 82.6 & \bf 83.3 & 95.8 & \bf 58.8 & \bf 65.0 & \bf 73.5 \\
    \bottomrule
\end{tabular}
}
\caption{Experiments on HotpotQA using FiD-Base as reader. We consider the settings illustrated in \autoref{fig:expsetting}. The best performance is highlighted in bold, while the second-best performance is underlined.}
\label{tbl:multi dataset experiment FiD-Base}
\end{table*}

\subsection{Multi-Hop Question Answering}\label{sec: Multi-Hop QA}
Multi-hop Question Answering (QA) systems typically follow a two-phase process: first, retrieving relevant passages, and then using these passages to answer the question. 2WikiHop~\cite{DBLP:conf/coling/HoNSA20}, HotpotQA~\cite{DBLP:conf/emnlp/Yang0ZBCSM18}, Musique~\cite{DBLP:journals/tacl/TrivediBKS22}, and MultiHop-RAG \cite{DBLP:journals/corr/abs-2401-15391} are widely used benchmarks for evaluating and improving RAG systems in handling complex multi-hop reasoning tasks. The retrieval strategies can differ depending on the QA setting, which may either be open-domain or reading comprehension. 

In open-domain QA, the focus is on retrieving relevant passages from a large corpus. Methods like MDR \cite{DBLP:conf/iclr/XiongLIDLWMY0KO21} and BeamDR \cite{DBLP:conf/naacl/ZhaoXBD21} are commonly used in this context. In the case of reading comprehension, retrieval methods are generally categorized into one-step and two-step approaches. One-step methods, such as SAE \cite{DBLP:conf/aaai/TuHW0HZ20}, rank passages by concatenating the question with each candidate passage. Two-step methods, including S2G \cite{DBLP:journals/corr/abs-2107-11823} and FE2H \cite{DBLP:conf/icassp/LiLY23}, start by selecting an initial hop passage and then refine the search by pairing it with additional candidates. The R$^3$ model \cite{DBLP:conf/cncl/YinWHWYZCHQ23} enhances this approach by selecting multiple passages at the outset and combining them to find the correct answer. Beam Retrieval \cite{DBLP:conf/naacl/ZhangZ00H24} further extends the process by using a beam search, enabling it to handle more complex multi-hop retrieval tasks that go beyond just two hops.

Our work focuses on a different scenario: we analyze the limitations of the LLM-Supervised Reranker Training strategy in widely used RAG systems for multi-hop question answering tasks (as detailed in \autoref{sec: Challenges in Handling Indirectly Relevant Documents with EMDR/PDist}) and propose an end-to-end reranker training strategy based on Gumbel Subset Sampling, which is well-suited for multi-hop question answering tasks.

\subsection{Dataset Description}\label{sec: appendix_datasets}
The datasets in our study encompass a variety of challenges designed to assess different facets of question answering. \textbf{HotpotQA}~\cite{DBLP:conf/emnlp/Yang0ZBCSM18} is a multi-hop QA dataset that requires reasoning over multiple Wikipedia articles to derive answers, emphasizing both factual retrieval and reasoning capabilities. Similarly, \textbf{2WikiHop}~\cite{DBLP:conf/coling/HoNSA20} extends the complexity of multi-hop reasoning by introducing questions that necessitate navigating a knowledge graph, enhancing the evaluation of entity-based information retrieval. \textbf{Musique}~\cite{DBLP:journals/tacl/TrivediBKS22} is designed to assess compositional reasoning by decomposing complex questions into a sequence of simpler sub-questions, providing a structured approach to multi-step reasoning evaluation. Meanwhile, \textbf{Natural Questions (NQ)}~\cite{DBLP:journals/tacl/KwiatkowskiPRCP19} presents real-world search queries answered using long-form documents, challenging models to extract and summarize information from extensive contexts. Lastly, \textbf{TextbookQA (TQA)}~\cite{DBLP:conf/acl/KimKK19} focuses on domain-specific comprehension, where questions require understanding of textbook-style knowledge, integrating both textual and diagrammatic content for a holistic assessment of contextual understanding and inferential capabilities.

\subsection{Additional Results with FiD-Base}
We present additional results on the HotpotQA dataset using FiD-Base as the reader, as shown in \autoref{tbl:multi dataset experiment FiD-Base}. Our method outperforms others across most metrics, demonstrating its efficacy.

\section{Challenges in Handling Indirectly Relevant Documents with EMDR/PDist}\label{sec: Challenges in Handling Indirectly Relevant Documents with EMDR/PDist}

Both EMDR$^2$~\cite{DBLP:conf/nips/SachanRHDY21, DBLP:conf/naacl/ShiMYS0LZY24, DBLP:conf/iclr/Lin0CSL00KSLZY24} and PDist~\cite{DBLP:journals/jmlr/IzacardLLHPSDJRG23, DBLP:conf/naacl/GlassRCNCG22} are based on the premise of distilling the importance of individual documents in a multi-document retrieval and generation process. While effective for ranking directly relevant documents, both methods encounter challenges when dealing with \textit{indirectly relevant documents}, which provide context but do not directly contain the answer. 

In EMDR$^2$, the objective is to maximize the log-likelihood of generating the answer given the query and individual documents. The loss function is given by:
\[
\mathcal{L}_{\text{EMDR}^2} = \log \left[ \sum_{k=1}^K p_{\textsc{lm}}(\mathbf{a} \ | \ \mathbf{q}, \mathbf{p}_k) p_{\mathcal{R}}(\mathbf{p}_k \ | \ \mathbf{q}) \right]
\]
where \( p_{\textsc{lm}}(\mathbf{a} \ | \ \mathbf{q}, \mathbf{p}_k) \) represents the language model’s probability of generating the answer conditioned on the query and document, and \( p_{\mathcal{R}}(\mathbf{p}_k \ | \ \mathbf{q}) \) is the reranker's preference for the \(k\)-th document. However, this approach assumes the independence of documents when generating the answer. Indirectly relevant documents, while crucial in providing context, do not appear to contribute meaningfully when evaluated independently. The importance of such documents can only be assessed when they interact with other documents in the evidence chain, making their relevance difficult to capture in this formulation.

Similarly, PDist minimizes the KL divergence between the reranker's distribution \(p_{\mathcal{R}}(\mathbf{p}_k \ | \ \mathbf{q})\) and the distribution \( p_k \) derived from the language model's perplexity. \( p_k \) represents the distribution over documents that the language model prefers based on its perplexity improvement:
\[
p_k = \frac{\exp(\log p_{\textsc{lm}} (\mathbf{a} \ | \ \mathbf{p}_k, \mathbf{q}))}{\sum_{i=1}^K \exp ( \log p_{\textsc{lm}} (\mathbf{a} \ | \ \mathbf{p}_i, \mathbf{q}))}
\]

In both methods, the language model \( p_{\textsc{lm}}(\mathbf{a} \ | \ \mathbf{q}, \mathbf{p}_k) \) computes the probability of generating the answer based on the query and a single document \( \mathbf{p}_k \), treating the document in isolation. This formulation assumes that each document, independently, provides enough information to determine the relevance to the query and the answer. However, in the case of indirectly relevant documents, this assumption breaks down. Indirectly relevant documents do not contain the answer directly but instead contribute to the reasoning process by supporting or contextualizing other documents. When evaluated alone, these documents may appear less relevant or even irrelevant, which undermines the effectiveness of both methods.

\begin{center}
\begin{algorithm*}
\caption{Learnable Sampling Weights Setting}
\label{alg:masking_no_reranker}
\begin{algorithmic}[1]
\Procedure{StochasticSubsetMask}{document weights $w_1, \dots, w_n$, temperature $\tau$, scale factor $\kappa$, subset size $k$}
    \For{$j = 1$ \textbf{to} $k$} 
        \State $\tilde{w}_i = -\log(-\log(u_i)) + \kappa \cdot w_i,\quad u_i \sim \mathcal{U}(0, 1) \quad \forall i \in [n]$
        \State $\hat{\mathcal{M}}^{j} = \max(\hat{\mathcal{M}}^{j-1}, \text{softmax}\left(\frac{(\tilde{w}_1, \dots, \tilde{w}_n)}{\tau}\right))$ \quad \textcolor{gray}{\# $\hat{\mathcal{M}}^{0} = [0, \dots, 0]$}
    \EndFor
    \State \Return $\hat{\mathcal{M}}^{k}$ \Comment{Return Relaxed top-$k$ Mask}
\EndProcedure
\State
\State Given a query $\mathbf{q}$, answer $\mathbf{a}$, and $n$ retrieved passages $\mathbf{p_1}, \dots, \mathbf{p_n}$
\State \textbf{Initialization:} Initialize learnable document weights $w_1 = 0, \, w_2 = 0, \, \dots, \, w_n = 0$
\For{each training step}
    \State $\hat{\mathcal{M}} = \operatorname{StochasticSubsetMask}(w_1, \dots, w_n, \tau, \kappa, k)$
    \State Apply $\mathcal{DMA}(\hat{\mathcal{M}})$ to obtain logits and language loss $\mathcal{L}_{LM}$  \Comment{\autoref{sec: DMA}}
    \State Update document weights $w_1, \dots, w_n$ with $\nabla_{w_1, \dots, w_n} \mathcal{L}_{LM}$ \Comment{Gradient-based update}
\EndFor
\end{algorithmic}
\end{algorithm*}
\end{center}

\begin{figure*}[!th]
  \includegraphics[width=0.9\linewidth]{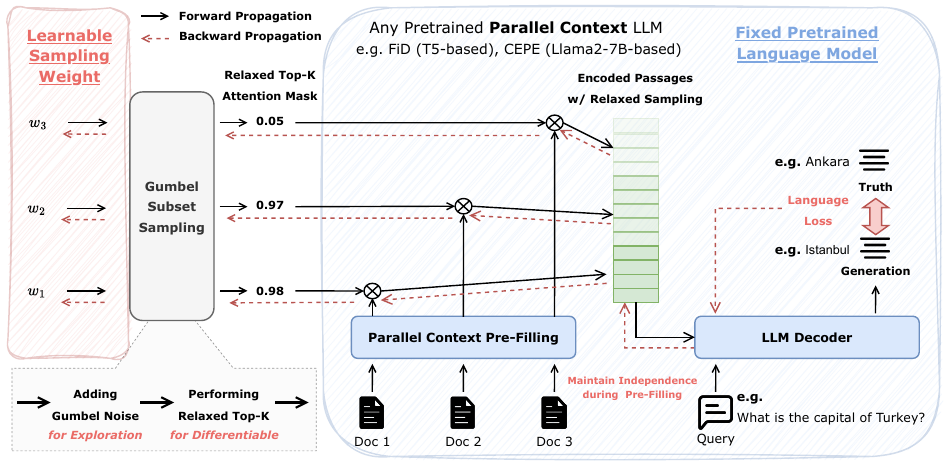}
  \centering
\caption{Setting of Learnable Sampling Weight. Optimizing candidate document sampling weights directly without leveraging reranker's prior textual knowledge.}
\label{fig:learnable_sampling_weight method}
\end{figure*}

\section{Irrelevant Document Setting}\label{sec: appendix_indirectly relevant documents setting}
Multi-hop question answering in \texttt{HotpotQA} involves synthesizing information from multiple documents to resolve a single query. Although these questions can be categorized into four major reasoning types---such as bridging intermediate information, comparing entities, verifying multiple properties, or inferring properties through a bridge---they all share the common requirement of gathering evidence across several sources. In this setting, identifying and assessing \emph{indirectly relevant documents} can be instrumental for measuring how effectively a model captures the full chain of reasoning. Our approach defines an \emph{indirectly relevant document} as any document labeled as relevant in the dataset yet not explicitly containing the final answer. This rule is rational in that it highlights the documents that contribute background or bridging information. However, this simple rule sometimes blurs the distinction between direct and indirect relevance. For instance, when a document only partially contains the answer, or when multiple sources each provide different fragments of a single reasoning chain (especially in question types like comparing entities or verifying multiple properties), all supporting documents will be classified as indirectly relevant by this rule. The concept of “partial” evidence is inherently difficult to categorize as either direct or indirect, and our rule consequently treats such “partial” evidence as indirectly relevant documents.

After processing the dataset, we observe that in the training set, the ratio of total queries to data entries that do contain \emph{indirectly relevant documents} is 90,447 to 664,247. In the development set, this ratio is 7,405 to 5,966. The relatively large number of data entries that do contain indirectly relevant documents allows for a robust evaluation of a model’s ability to retrieve and utilize such supporting evidence. Thus, this formulation not only aligns well with the structure of \texttt{HotpotQA} but also provides a meaningful benchmark for analyzing the effectiveness of different methods in capturing multi-hop dependencies beyond direct answer retrieval.

\section{Learnable Sampling Weights Setting}\label{sec: appendix_Learnable Sampling Weights}
\subsection{Scalar Relevance Baselines}
In our thesis, we employ different scalar metrics to quantify the relevance of each candidate document in the RAG system. These metrics correspond to the different LLM-supervised reranker training methods, and they serve as proxies for the document's contribution to generating the correct answer. In particular, we consider the following three metrics:

\paragraph{Lowest Perplexity}
For methods such as EMDR$^2$ and Perplexity Distillation (PDist), each candidate document $\mathbf{p}_k$ is evaluated by combining it with the query $\mathbf{q}$ and computing the language model’s negative log-likelihood of generating the ground-truth answer $\mathbf{a}$. Formally, the scalar relevance score is defined as:
\[
s_{\text{perplexity}}(\mathbf{p}_k) = -\log p_{\textsc{lm}}(\mathbf{a} \mid \mathbf{q}, \mathbf{p}_k).
\]
In this setting, a lower perplexity (i.e., a higher value of $s_{\text{perplexity}}$) indicates that the document better facilitates the generation of the correct answer, and is therefore considered more relevant.

\paragraph{Highest Attention Score}
The Attention Distillation (ADist) method evaluates relevance by feeding all candidate documents to the language model simultaneously and aggregating the cross-attention scores. For each document $\mathbf{p}_k$, the relevance score is computed by weighting the attention score $\alpha_k$ with the L2 norm of its corresponding value vector $\mathbf{v}_k$:
\[
s_{\text{attn}}(\mathbf{p}_k) = \alpha_k \,\|\mathbf{v}_k\|_2.
\]
Here, a higher attention score signifies that the language model assigns more importance to the document during answer generation, thereby indicating greater relevance.

\paragraph{Highest Leave-one-out Perplexity}
The Leave-one-out Perplexity Distillation (LOOP) method assesses the impact of each candidate document by measuring the degradation in the language model’s performance when that document is excluded from the candidate set. For each document $\mathbf{p}_k$, the relevance score is defined as:
\[
s_{\text{loop}}(\mathbf{p}_k) = -\log p_{\textsc{lm}}\Big(\mathbf{a} \mid \mathcal{D}_K \setminus \{\mathbf{p}_k\}, \mathbf{q}\Big),
\]
where $\mathcal{D}_K$ denotes the set of top-$k$ candidate documents. A higher leave-one-out score implies that the removal of $\mathbf{p}_k$ leads to a significant deterioration in the language model’s ability to generate the answer, marking it as highly relevant.

\subsection{Our method}
In our proposed approach, we eliminate the dependency on a dedicated reranker by replacing it with a set of learnable scalar weights—one per document—that are initialized at zero and updated directly via gradients from the language modeling loss computed by the differentiable masked attention module. As detailed in Algorithm~\ref{alg:masking_no_reranker} and \autoref{fig:learnable_sampling_weight method}, the algorithm employs stochastic Gumbel noise to perform a relaxed top‑$k$ selection over these weights, ensuring that the entire process remains fully differentiable. This method iteratively refines the document weights over multiple steps on a single query–answer pair and its corresponding documents, thereby enabling the model to learn which documents are most informative for the downstream language modeling task.

\begin{table*}[!th]
\small
\centering
\begin{tabular}{p{5cm} p{5cm} p{4cm}}
\toprule
\textbf{Dataset} & \textbf{URL} & \textbf{License} \\
\midrule
\textbf{Multi-hop QA} & & \\
2WikiMultiHopQA~\cite{DBLP:conf/coling/HoNSA20} & \url{https://github.com/Alab-NII/2wikimultihop} & Apache License 2.0: \url{https://github.com/Alab-NII/2wikimultihop/blob/master/LICENSE} \\
HotpotQA~\cite{DBLP:conf/emnlp/Yang0ZBCSM18} & \url{https://hotpotqa.github.io/} & CC BY-SA 4.0: \url{https://hotpotqa.github.io/} \\
MuSiQue~\cite{DBLP:journals/tacl/TrivediBKS22} & \url{https://github.com/stonybrooknlp/musique} & CC BY 4.0: \url{https://github.com/stonybrooknlp/musique/blob/main/LICENSE} \\
\midrule
\textbf{Single-hop QA} & & \\
Natural Questions (NQ)~\cite{DBLP:journals/tacl/KwiatkowskiPRCP19} & \url{https://ai.google.com/research/NaturalQuestions} & CC BY-SA 3.0: \url{https://ai.google.com/research/NaturalQuestions/download} \\
Textbook Question Answering (TQA)~\cite{DBLP:conf/acl/KimKK19} & \url{https://prior.allenai.org/projects/tqa} & CC BY-NC 3.0: \url{https://prior.allenai.org/projects/tqa} \\
\bottomrule
\end{tabular}
\caption{Summary of URLs and Licenses for Datasets}
\label{tab:qa_datasets}
\end{table*}

\section{URLs and Licenses}
\autoref{tab:qa_datasets} provides license information for the datasets we utilize in our experiments. We employ all the above datasets solely for research purposes, in accordance with their designated uses.

\end{document}